\documentclass[sigconf,balance=false]{acmart}

\usepackage{amsfonts}

\usepackage[utf8]{inputenc}%
\usepackage{graphicx}
\usepackage{caption}
\usepackage{subcaption}
\usepackage{color}
\newcommand{\revShort}[2]{#2}
\newcommand{\revCut}[1]{}

\usepackage[normalem]{ulem}

\newcommand{\cutAA}[1]{}
\newcommand{\revAA}[2]{#2}
\newcommand{\revBB}[2]{#2}
\newcommand{\revCC}[2]{#2}

\newcommand{\revDD}[2]{#2}

\newcommand{\revEE}[2]{#2}
\newcommand{\commentout}[1]{}

 \usepackage{popets}
	\setcopyright{popets}
 	\copyrightyear{2023}
	\acmYear{2023}
	\acmVolume{2023}
	\acmNumber{1}
	\acmDOI{XXXXXXX.XXXXXXX}
	\acmISBN{}
	\acmConference{Proceedings on Privacy Enhancing Technologies}
\begin{document}

\author{Minh-Ha Le}
\affiliation{%
  \institution{Link\"oping University} %
  \country{Sweden}}

\author{Niklas Carlsson}
\affiliation{%
  \institution{Link\"oping University}
  \country{Sweden}}

  \title{StyleID: Identity Disentanglement for Anonymizing Faces}

  \begin{abstract}
  Privacy of machine learning models is one of the remaining challenges that hinder the broad adoption of Artificial Intelligent (AI). This paper considers this problem in the context of image datasets containing faces.  Anonymization of such datasets 
  \revAA{are}{is} 
  becoming increasingly important due to their central role in the training of autonomous cars, for example, and the vast amount of data generated by surveillance systems. While most prior work
  \revAA{de-identify}{de-identifies} 
  facial images by modifying identity features in pixel space, we instead project the image onto 
  \revAA{}{the}
  latent space of a Generative Adversarial Network (GAN) model, find the features that 
  \revAA{provides}{provide} 
  the biggest identity disentanglement, and then manipulate these features in latent space, pixel space, or both.  
  The main contribution of the paper is the design of
  a feature-preserving 
  anonymization 
  \revAA{framework that}{framework, StyleID, which} 
  protects the individuals’ identity, while preserving as many characteristics of the original faces in the image dataset as possible.  As part of 
the contribution, we present a novel
  disentanglement metric, three complementing disentanglement methods, and new insights into identity disentanglement. \revAA{The framework}{StyleID}
  provides tunable privacy, has low computational complexity, and is shown to outperform current state-of-the-art solutions.
\end{abstract}

\keywords{Identity disentanglement, anonymization, feature-preserving, privacy, StyleGAN, face editing}

\maketitle
\section{Introduction}

Machine learning is currently considered one of the primary enablers for future technological advancements.  
\revDD{Since the quality of machine learning models to a large extent 
\revAA{depend}{depends} 
on the quality and 
\revAA{the size}{size} 
of the datasets 
\revAA{}{that}
they are trained on, any AI technology operating in 
environments where humans are present
\revAA{therefore need}{needs} 
access to large datasets that contain images of humans.}{
AI technologies operating in environments including humans are therefore expected to need access to large datasets containing images of humans.}

However, large image datasets containing real faces quickly cause privacy concerns.  For example, the police’s use of ClearView AI to track criminals over the world has received significant media spotlight~\cite{hill2020secretive}.
With 3 billion images of people in the 
\revAA{database (a number equal to close to half the world's population),}{database,} 
concerns have been raised regarding who is in the database and the police’s right to use it 
\revAA{as a surveillance tool.}{for surveillance.}
The privacy risks of such image databases become even greater if considering the potential consequences of others using 
\revAA{such}{the} 
databases 
for their own 
\revAA{(greedy) benefit.}{benefit.}

Here, it is important to note that it typically is not the AI technologies themselves that present the privacy risks; it is the datasets, and how the datasets and the models 
\revAA{that they enable}{created using these datasets} 
are being used.  To prevent misuse of the datasets while at the same time 
\revAA{enable}{enabling} 
the development of future applications in a privacy-conscious manner, it is therefore important that datasets
\revBB{easily can}{can}  
be anonymized in ways that preserve the utility of the datasets.  If done well, such 
tools (and the generated datasets) 
\revAA{can provide a 
silver bullet
that can}{will}
benefit machine learning technologies 
\revBB{that}{(that} 
require training using datasets containing images of 
\revAA{humans.}{humans) and other fields (e.g., image editing, synthetic 2D/3D-avatars, 
privacy on social media).}

Anonymization can easily be done using occlusion and confusion methods that hide the identity of the faces. However, such methods typically significantly reduce the utility of the datasets and the accuracy that can be expected by the machine learning models trained on such data.  Anonymizing facial datasets in ways that preserve the facial characteristics of both the individual images and the dataset as a whole is a much harder task.  
\revAA{On}{One}
reason for this is that facial images represent one of the most complex information types and faces provide 
\revAA{}{a}
direct identity representation of humans.  This is perhaps why most prior %
work
primarily 
\revAA{have}{has}
focused on the naturalness of the anonymized faces~\cite{le2020anonfaces, gafni2019live, hukkelaas2019deepprivacy, maximov2020ciagan} or proved basic properties such as k-anonymity~\cite{newton2005preserving, gross2006model}.  

In this paper, we take a more ambitious approach in which we disentangle and hide the identity-related facial features while aiming to preserve the main visual characteristics of the faces.  
\revAA{For this,}{To achieve this objective,} 
we present novel identity disentanglement approaches that operate in latent space, pixel space, or both.    
The solutions presented are part of 
\revAA{a framework}{our framework, called StyleID,} 
that (1) 
\revAA{identify and manipulate}{identifies and manipulates} 
the identity-relevant information in a face 
\revAA{so to}{to} 
provide an anonymized face, 
while (2) preserving non-identity-related features (e.g., pose, facial expression, background, and hair), (3) without destroying the facial naturalness. As of today, several papers have shown how GANs can be used for one or two of these aspects at a time. (See Sec.~\ref{sec:related}.)
However, to our knowledge, we are 
\revAA{}{the}
first to address all three aspects simultaneously.

Much of the de-identification works do not preserve attributes and/or lack in naturalness~\cite{wu2019privacy, li2019anonymousnet, gafni2019live}. Other works use pre-trained image generators such as StyleGAN~\cite{karras2019style, karras2019analyzing} to achieve high naturalness, but often only focus on the manipulation of attributes in the facial images~\cite{shen2020interfacegan, harkonen2020ganspace, patashnik2021styleclip, wu2021stylespace}; not de-identification.
Overall, there is very limited work studying the privacy-sensitive information in both latent space and pixel space. 

\revAA{{\bf Main contributions:}}{{\bf Contributions:}}
The main contribution of this work is the design of a {\em feature-preserving anonymization framework} that uses our new approach to protect the individuals’ identity, while preserving as many characteristics of the original faces in the image dataset as possible.
However, in the derivation of this design, the paper makes several additional important contributions. 
First, at the core of the design are three novel methods for {\em identity disentanglement}.  The methods are complementing each other, 
\revAA{builds}{build}
upon each other, and operate either in latent space (only) or in both latent and pixel space simultaneously.  For example, the first method identifies and manipulates the part of the latent space that provides the most disentanglement, the second method uses segmentation masks together with insights from the first method to generate random faces for which we control the overlap in pixel space, and the third method builds a model that automates the disentanglement (and anonymization) process that operates in latent space but is trained using a ground truth built upon the second method.  

Second, we present a novel identity disentanglement metric that we use to derive insights, evaluate methods, and demonstrate that effective identity disentanglement is possible in latent space.  The results of our evaluations
\revAA{provides}{provide} 
new insights into how to best hide privacy-sensitive information in both latent and pixel space.  

Third, we use these insights to incorporate the methods into 
\revAA{the}{our StylelID}
framework so 
\revAA{}{as}
to provide tunable
\revAA{anonymity, allowing}{anonymity and attribute preservation tradeoffs.  
The framework allows} 
us to transform the identity-relevant information in a face 
\revAA{to}{into} 
an anonymized face with 
\revAA{}{a}
desirable level of anonymity, while preserving identity-irrelevant features 
\revAA{(e.g., pose, facial expression, hair, and background) and}{and} 
the naturalness.  Furthermore, the methods 
\revAA{have relatively 
low computational cost}{are efficient}
(e.g., the first two methods only use pre-trained models) and 
\revAA{are shown to outperform}{outperform} 
current state-of-the-art anonymization solutions.

\revCut{
{\bf High-level anonymization approach:}
The basic idea of our design is to first perform identity disentanglement in latent space and then apply changes to the aspects of a face (in latent space, pixel space, or both) that provide the most attractive tradeoffs between anonymization and preserving (non-identity revealing) facial features.
This idea is inspired by prior works that use StyleGAN to manipulate various attributes in facial images (e.g., the smile, aging, gender, etc.~\cite{shen2020interfacegan, harkonen2020ganspace, patashnik2021styleclip, wu2021stylespace}).  However, rather than identifying and manipulating such easily observed attributes, we explore how to best disentangle the identity within latent space and then use this knowledge to manipulate only the most identity-revealing information, while keeping the other image and face attributes intact.
This is a much harder problem since the identity typically can not be attributed to an easily identifiable attribute in an image.

In contrast to the case of attribute disentanglement, where visible attributes often can be mapped onto a clear latent vector (e.g., onto a scale that measures happiness, eye-openness, hair length, age, etc.) that can be validated by the human eye, identity disentanglement is much more subtle.  To address this problem we build a latent mask using pairs of training images that satisfy the desired requirements.
To do so we propose a metric to measure the identity disentanglement of two faces, and then use the metric to identify the parts of the latent space that provide the most attractive tradeoffs between hiding identity-related features and impacting the non-identity-related attributes.
Using one of several tunable objective function, the three anonymization methods associated with each identity disentanglement method then (1) manipulates the latent codes of an image, (2) swaps some masked region of a face with that of a carefully generated face, or (3) uses a feature-preserving swapper, respectively, so to move the identity further from the original face or towards a (randomly generated) target identity. 

Finally, we note that this is the first work attempting to find the identity direction in the latent space. Compared to the work of \cite{nitzan2020face}, which only disentangle identity from pose and facial expressions, our work achieves a higher level of identity disentanglement both in latent space and in pixel space. 
Leveraging the power of StyleGAN also helps us achieve more natural looking results than them and other GAN-based models.  
With our approach, facial anonymization is as simple as changing face attributes such as smile, age, gender, etc. Furthermore, disentangling latent codes in the latent space gives us several other advantages, including high efficiency, tunability, and flexibility. 
For example, we do not have to re-train our GAN models (known to be expensive both in terms of computation and complexity) and we can tune the privacy by controlling how far the generated face is from the original face and which privacy-sensitive attributes are included.  

}%

{\bf Outline:}
Sec.~\ref{sec:framework} presents an overview of the framework, our disentanglement methods, and defines our disentanglement metric.  Secs.~\ref{sec:layers} and~\ref{sec:channels} 
\revDD{then present}{present} 
how 
\revAA{identity disentanglement in latent space can be used for anonymization}{\revDD{anonymization through the use of identity disentanglement in latent space can be achieved}{anonymization can be achieved}} 
by manipulating layers or channels of the latent codes associated with a face.  Our feature-aware identity masking method (Sec.~\ref{sec:mask}) and the latent swapper (Sec.~\ref{sec:mapper}) are presented 
\revDD{in the following sections.}{next.}  Sec.~\ref{sec:evaluation} 
\revDD{then evaluates}{evaluates}
\revAA{the framework (and methods)}{StyleID (and its methods)} 
against facial recognition tools, based on how well they preserve attributes and the identity diversity they provide.  
Finally, Sec.~\ref{sec:related} 
\revDD{presents a comparison against}{compares with} 
related 
\revAA{works,}{works, Sec.~\ref{sec:broader} discusses security and ethical considerations,} 
\revDD{before}{and} 
Sec.~\ref{sec:conclusions} presents our conclusions. 

\section{Framework overview}\label{sec:framework}

This paper focuses on the {\em anonymization} of
\revAA{}{image}
datasets including faces.
In addition to removing identifying characteristics so to protect the identity of individuals (i.e., de-identification) and 
\revAA{ensure}{ensuring} 
that the faces are altered in such a
way that each face no longer can be related back to a given individual (i.e., the process should not be revertible), a good anonymization technique should ensure high utility of the resulting data.  

To ensure high utility, 
we present the design of 
\revAA{}{StyleID,} 
a {\em feature-preserving anonymization framework} that
transforms the identity-relevant information in a face 
\revAA{to}{into} 
an anonymized 
\revAA{face, which provides}{face of} 
the desirable level of anonymity, while preserving as many non-identity-related features (e.g., pose, facial expression, image background, and hair) as possible 
\revBB{without and}{and} 
maintaining the face's naturalness. 

At the core of the design 
\revAA{are}{is} 
the basic idea of first applying identity disentanglement in latent space and then applying changes to the aspects of a face (in latent space, pixel space, or both) that provide the most attractive tradeoffs between anonymization and 
feature preservation.
\revAA{}{For much of this manipulation we leverage the latent space of StyleGAN~\cite{karras2019style,viazovetskyi2020stylegan2}.  In addition to allowing facial editing in latent space, StyleGAN has been found (by human evaluators) to produce more realistic/natural images than other state-of-the-art solutions~\cite{zhou2019hype}. With the naturalness of the generated images of any framework being bounded by the quality of the face generator used, our choice to use StyleGAN's 
generator allows us to achieve higher naturalness than most prior anonymity frameworks~\cite{hukkelaas2019deepprivacy,maximov2020ciagan,newton2005preserving,shan2020fawkes}.} 

\subsection{Identity disentanglement approaches}\label{sec:approaches}

\revDD{To achieve 
\revAA{the above}{our} 
objectives, we}{We} 
present three complementing 
\revDD{identity disentanglement}{disentanglement}
approaches and show how
\revDD{they can be used for anonymization.}{to use them to achieve our anonymization objectives.} 

{\bf Disentanglement in latent space (Secs.~\ref{sec:layers} and~\ref{sec:channels}):}
The first approach operates in latent space.  With this approach, we (1) project the face images into latent space, (2) identify the layers/channels in the latent code that contribute the most to the individual’s identity, (3) manipulate these layers/channels of the latent code in a desirable direction (e.g., as far away from the 
\revAA{origin}{original source} 
face or towards an alternative target identity), before (4) generating the final image.   For tunable privacy, we control how far away from the 
\revAA{original source}{source} 
identity we push the identity.  
\revAA{Here, we consider both the case where we push the latent codes in a direction orthogonal to the original identity or in the direction of either an existing face or a randomly generated face.  While}{While} 
this approach performs well, 
it is not able to (on its own) consistently preserve the background and some facial features (e.g., hair, facial expression, eyes direction) and may create problems when applied 
\revAA{on}{to} 
video.

\begin{figure}[t]
\centering
\includegraphics[width=0.86\linewidth]{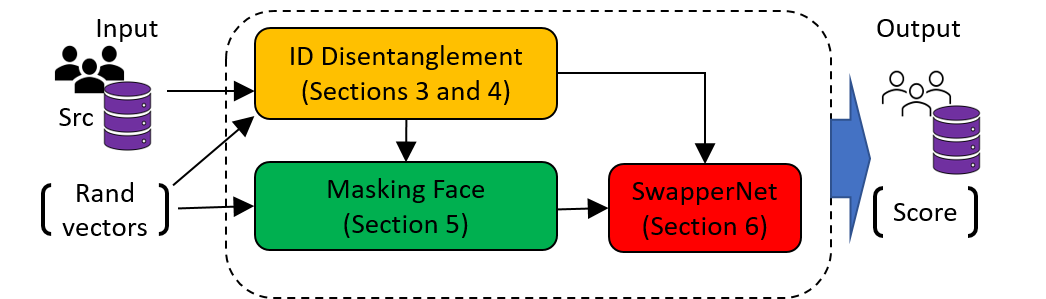}
\vspace{-10pt}
\caption{Overview of our 
\revAA{}{StyleID} 
framework.}
\label{fig:overview}
\vspace{-12pt}
\end{figure}

{\bf Disentanglement in pixel space (Sec.~\ref{sec:mask}):}
We then present a technique that incorporates the use of {\em segmentation masks} to improve the quality of the generated images and to better preserve selected features as seen in pixel space. This approach is attractive for videos and other contexts that place stricter requirements on preserving specific facial features and/or the background. Using an example implementation, we demonstrate how the use of segmentation masks combined with some of the insights from our first approach can be used to generate random faces that have the same face mask and how this can significantly simplify accurate face swapping. The approach allows us to effectively generate faces with a matching pose, provides fine-grained control of 
\revAA{the how}{how} 
much of the original identity (and face) is preserved in the generated output image and is relatively lightweight.  While this approach solves some of the problems of the first approach, we have found that it can result in a mismatch of lighting between the randomly generated face and the original face.  Although such issues in most (but not all) cases can be nicely corrected by a match color process, we note that such a process requires additional care.

{\bf Latent swapper (Sec.~\ref{sec:mapper}):}
Finally, to address the remaining shortcomings, we build a model that automates the process of anonymizing a face in latent space with a ground truth built 
\revAA{upon using}{using} 
our segmentation mask approach. 
\revAA{To do this, the}{The} 
model finds an $\alpha$-mask and trains swapper modules 
\revAA{that are weighted}{weighted} 
based on the insights derived from our 
\revAA{basic latent}{latent} 
space approach to hide the source identity beyond the identification threshold of modern  facial recognition systems 
\revAA{}{(FRS:s)}
while limiting the changes to non-identity related attributes. 

Fig.~\ref{fig:overview} presents an overview of
\revAA{the disentanglement methods that we have 
implemented for the three approaches and how we use them for anonymization.}{StyleID and the three disentanglement methods designed and implemented within the framework.}
\revAA{}{(1) The {\em ID Disentanglement} component (yellow) is at the heart of our design.  It is used to disentangle identity information in latent space. 
(2) 
\revAA{The {\em Masking Face} component}{{\em Masking Face}}
(green) is applied in pixel space. This component is used to generate and apply random identity masks.
(3) {\em SwapperNet} is an all-in-one anonymizer model that swapps the original identity to a random identity.
While each of the three components builds on the prior component(s), 
\revAA{we note that each}{each} 
component can produce anonymized faces also without the use of the later component(s).}

\subsection{\revAA{}{Privacy-utility tradeoff}}

\revAA{}{The main problem of anonymizing datasets is 
finding
a desirable tradeoff between privacy and utility. One challenge 
\revDD{to this 
key problem is}{is} 
the lack of universal metrics to quantify the tradeoff.  Here, we present 
and motivate 
the 
primary
privacy and utility metrics used in this work.} 

\revAA{}{First, motivated by the importance of protecting the facial identity against 
\revAA{facial recognition system (FRS),}{FRS:s,} 
we use the identity distance 
calculated using
popular face embedding models used by such systems to measure privacy. Ideally, an anonymized face $o$ should have an identity distance to the original 
source
face that exceeds some 
\revDD{reasonable threshold.}{threshold or is considered ``far-enough" away from the source.} 
Second, we measure the utility using attribute scores extracted from both the source faces and the generated output faces.
We next formalize the above metrics of privacy and utility, as calculated over a full dataset.}

\revBB{}{{\bf Definition (Privacy metrics):} 
Let $\mathcal{A}$ be an anonymizer converting a source dataset $\mathbb{S}$ containing $N$ human faces $\{S_1, S_2, ..., S_N\}$ into an anonymized output dataset $\mathbb{O}$ containing $N$ output faces $\{O_1, O_2, ..., O_N\}$. Furthermore, let $IdNet$ be a face embedding model.  
Now, the identity distances between a given source face $S_i$ and the corresponding output face $O_i$ can be calculated as follows:
\begin{align}
\Delta_i^{ID} & = \delta(IdNet(S_i), IdNet(O_i)),
\end{align}
where $\delta()$ is the pairwise distance calculated over the embedding vectors $e^S_i = IdNet(S_i)$ and $e^O_i = IdNet(O_i)$, for $1 \leq i \leq N$. Using this distance, we define the privacy metric $I$ as the average identity distances of all the source-output pairs; i.e., $I = \frac{1}{N} \sum_{i=1}^N \Delta_i^{ID}$.} 

\revAA{}{For the purpose of evaluation, we also report the probability that an arbitrary pair satisfies a privacy threshold $\Gamma$ and consider the anonymizer $\mathcal{A}$ to ``strictly" satisfying a privacy guarantee $\Gamma$ if $\Delta_i > \Gamma$, $\forall O_i \in \mathbb{O}$). We also calculate the ROC curve and report the average rank of the output image $O_i$ when comparing the output images distances to the source image $S_i$.}

\revDD{}{The above metric and statistics are based on the decision process of current state-of-the-art FRS:s, not on probability theory. Future work could involve the use of a more formal privacy definition.}

\revAA{}{{\bf Definition (Utility metrics):} Again, consider source set $\mathbb{S}$, an anonymizer $\mathcal{A}$, and a corresponding output set $\mathbb{O}$.  Furthermore, let $AttrNet$ be a face attribute classifier model that extracts $M$ attributes $\{a_1, a_2, ..., a_M\}$ from the face, where $a_i \in (0,1)$, $1 \leq i \leq M$. Now, the attribute distance between a source-output pair $i$ ($1 \leq i \leq N$) can be calculated as: 
\begin{align}
\Delta_i^{Attr} = \delta(AttrNet(S_i), AttrNet(O_i)),  
\end{align}
where $\delta()$ is the pairwise distance calculated over the attribute vectors $a^S_i = AttrNet(S_i)$ and $a^O_i = AttrNet(O_i)$.
Using this distance, we define the utility metric $A$ as the average attribute distances of all corresponding pair source-output faces; i.e., $A = \frac{1}{N} \sum_{i=1}^N \Delta_i^{Attr}$.}

\revAA{}{For evaluating utility, we also measure the anonymizer's ability to preserve the distribution of the attributes observed in the dataset and the identity diversity.}

\subsection{Identity disentanglement metric}

Identity disentanglement in latent space has not been addressed by prior work and is not possible with existing methods~\cite{harkonen2020ganspace, wu2021stylespace, patashnik2021styleclip}.  
\revAA{One of the main challenges}{One main challenge} 
is that the identity is a complex combination of several facial features. 
\revAA{To allow us to}{To} 
perform identity disentanglement, we first define a metric to measure the
\revAA{degree of identity disentanglement.}{identity disentanglement achieved by an anonymizer $\mathcal{A}$.} 
Ideally, the metric should be sensitive to 
\revAA{changes in the identity}{identity changes} 
and insensitive to 
\revAA{changes in 
\revAA{non-identity}{non-identity-related} 
attributes.}{non-identity-related changes.}

To achieve this goal,
\revAA{our metric combines the}{we try to simultaneously maximize the identity distance and minimize the attribute distance.  For this reason, our metric combines the}
\revAA{results obtained using two pre-trained models: $IdNet$ and $AttrNet$. 
The $IdNet$ model calculates the embedding vectors $\overline{e_S}$ and $\overline{e_O}$ of the source face $S$ and output face $O$, respectively. 
Similarly, the $AttrNet$ model calculates attribute vectors $\overline{a_S}$ and $\overline{a_O}$ of the corresponding faces.}{privacy and utility metrics from the previous 
subsection.}
\revAA{Using these outputs, we then measure the scoring change going from face $S$ to $O$ as:
\begin{align} \label{eqn:score}
    IA_{score}(S,\revBB{G}{O}) & = \alpha \cdot h(\delta(IdNet(S), IdNet(\revBB{G}{O})) \\
    & ~~~ - \beta \cdot h(\delta(AttrNet(S), AttrNet(\revBB{G}{O})), \nonumber
\end{align}}{In particular, we calculate the disentanglement score going from face $S_i$ to $O_i$ as follows:
\begin{align} \label{eqn:score}
    IA_{score}(S_i,O_i) & = \alpha \cdot h(\Delta_i^{ID}) - \beta \cdot h(\Delta_i^{Attr}),
\end{align}}
where $\alpha$ and $\beta$ are 
\revAA{}{tunable}
constants, \revAA{$\delta()$ calculates the pairwise distance, and}{and} 
$h()$ is a normalization 
\revAA{function. In our case,
\revAA{we use the Euclidean distance 
\revBB{(i.e., $\delta(x,y) = \sqrt{(x_1 - y_1)^2 + ... + (x_n-y_n)^2}$)}{} 
and normalize}{we normalize} 
the identity and attribute distances into the same value range using min-max 
\revAA{scaling (i.e., $h(x) = \frac{x-X_{\min}}{X_{\max} - X_{\min}}$, where $X_{\min}$ and $X_{\max}$ are the minimum and maximum values observed).}{scaling.}}{function (max-min scaling) that normalizes identity and attribute distances into the same value range.}  
Through careful selection of $\alpha$ and $\beta$ we have found \revAA{that this}{the} metric 
\revAA{can}{to} 
nicely capture how successful a change from face 
\revAA{$S$ to $O$}{$S_i$ to $O_i$} 
was 
\revAA{in}{at} 
modifying the identity 
\revAA{of the face without}{without} 
significantly changing 
\revAA{the}{the facial} 
attributes.
\revCC{}{For the experiments, we use $\alpha$$=$$1$, $\beta$$=$$1.25$, and the Euclidean pairwise distance $\delta$.}

{\bf Implementation details:}
We use several pre-trained models in our experiments.
The image generator $G$ used is a StyleGAN2 model~\cite{viazovetskyi2020stylegan2} trained on the FFHQ dataset~\cite{karras2019style} at 1024x1024 resolution.  We use pSp~\cite{richardson2021encoding} 
\revAA{as our image encoder, which itself has been trained on the same dataset.}{(trained on the same dataset) as our image encoder.}
We use the state-of-the-art facial recognition models CurricularFace~\cite{huang2020curricularface} and ArcFace~\cite{deng2018arcface} to calculate the identity embeddings of $IdNet$ 
\revAA{(that we use to calculate identity distances between faces).}{(used to calculate identity distances).} 
Finally, the attribute predictor $AttrNet$ used is a custom MobileNet model~\cite{howard2017mobilenets} that is trained on the CelebA dataset~\cite{liu2015faceattributes} to predict 40 binary attributes. The predictor outputs a confidence vector in the value range (0,1).  \revBB{To emphasize the change in attribute values, we apply a logit function on each attribute value (i.e., $\textrm{logit}(x) = \log(\frac{x}{1-x})$) before forming the two attribute vectors $\overline{a_S}$ and $\overline{a_O}$ (that we then calculate the distance between).}{}

\subsection{Tunable anonymity}

All our disentanglement methods 
are designed to allow tunable 
\revDD{anonymity. We next define what we mean with tunable anonymity and the levels of privacy targeted.}{anonymity (define next) for the levels of privacy targeted.}

{\bf Definition (Tunable anonymity):} 
{\em Given a source face 
\revBB{an anonymizer (with adjustable privacy/utility parameters)}{$\mathbb{S}$, an anonymizer $\mathcal{A}$} 
should be able to adjust the level of privacy 
\revAA{that it offers}{offered} 
based on the parameters assigned for the specific 
\revAA{use case}{use-case} 
scenario.}  

Here, we consider and target three levels of privacy:
\begin{itemize}
\item[-]
{\em Low:} Small modification to identity features \revBB{}{$I \approx \Gamma$, where $\Gamma$ is the detection threshold of the FRS,} that are sufficient to fool FRS but that are barely noticeable to the human eye (e.g., privacy filters that appear like noise to a user). \revBB{}{In this case, ideally all $M$ attributes from the source 
$\{a_0, a_1, ..., a_M\}$ are preserved, meaning that $A$ is greater than a threshold $\Theta$ or $AttrNet$ successfully extracts all attributes $a_i > \theta$, $\forall O \in \mathbb{O}$, where 
$\theta$ is 
the binary decision threshold of $AttrNet$.}
\item[-]
{\em Medium:}
\revCC{Modifications to}{Preserving} a subset of facial features $\mathbf{Q} \in \mathbf{M}$ that result in identity changes both to an FRS \revBB{and humans}{$I \approx \Gamma$ and humans, i.e., 
$\forall O \in \mathbb{O}$, there exist\revCC{}{s} $ a_i \revCC{<}{>} \theta$, $\forall a_i \in \mathbf{Q}$ }.
\item[-]
{\em High:}  
\revAA{At this level, we also}{One may also want to} 
enforce 
\revAA{l-diversity and t-closeness}{desirable attribute distribution properties among the set of generated faces (e.g., t-closeness and l-diversity~\cite{li2007t}).}
\revAA{when considering facial attributes such as hair, ears, expression, and skin color}{This is to limit the individual-specific information 
\revDD{an observer can learn from observing the attributes of each face.}{revealed by the observed face attributes.}}
\end{itemize}

\revAA{}{For each of the three proposed disentanglement approaches, 
\revDD{the paper}{we}
demonstrates how to tune the level of anonymization among the first two levels (``low" and ``medium"). To achieve the highest level of anonymization, one can
add an outer loop on-top of our framework so as to ensure that 
some desirable
attribute-distribution properties (e.g., l-diversity,
t-closeness) are satisfied.
However, such extensions 
\revDD{(and the evaluation therefore) are}{are} 
considered outside the scope of this paper.  Here, we just note that our ability to control selected attributes in the 
\revDD{generation of output}{generated} 
images enables also this level of anonymization (if desired).}

\section{Identity disentanglement and swapping layers in latent space}\label{sec:layers}

\subsection{Background: StyleGAN + latent space}

StyleGAN is a generative model capable of generating realistic images from random noise vectors. 
With StyleGAN, the input latent codes $z \in Z$ are passed through 
\revCC{the encoder}{a mapper}, 
which 
\revCC{through}{is} 
a sequence of fully connected layers 
\revCC{outputs}{outputting} 
intermediate latent codes $L \in W$.  
\revAA{In StyleGAN, a latent code $L \in W$ is a two-dimensional array}{Here,
$L$
is a two-dimensional array} 
of size 18 $\times$ 512, where 
\revAA{each row is called a layer}{rows are called layers} 
and the values in each layer are called channels. 

One attractive property of StyleGAN is that its intermediate latent space $W$ is highly disentangled. For example, Karras et al.~\cite{karras2019style} observed that certain layers 
\revAA{in the latent space}{in} 
$W$ correspond to specific subsets of facial features and that the layers can be split into three categories:
\revAA{The {\em course layers}}{{\em Course layers}} (0 to 3) represent high-level attributes such as pose, face shape, hairstyle, and 
\revAA{eyeglasses, the {\em middle layers}}{eyeglasses. {\em Middle layers}}
(4 to 8) represent features such as the structure of the eyes, mouth, and 
\revAA{nose, and, finally the {\em fine layers}}{nose.  {\em Fine layers}} 
(8 to 17) hold the color scheme and micro-structure.

Prior work has shown how such disentanglement can be leveraged to manipulate selected facial features~\cite{harkonen2020ganspace, shen2020interfacegan, wu2021stylespace, patashnik2021styleclip}. By carefully manipulating the latent codes these works have successfully changed one feature at the time without changing the identity of the face.  However, thus far no work has considered the opposite problem; i.e., how to 
change the identity while preserving other facial features.  In this section, 
\revAA{we analyze to what degree such disentanglement is possible.}{we consider such disentanglement.}

\vspace{-2pt}

\subsection{High-level approach and key problems}

\revDD{As outlined in \revBB{Section}{Sec.}~\ref{sec:approaches}, our approach for disentanglement in latent space includes four key 
\revShort{steps:  (1) projecting face images onto latent space, (2) identifying the layers (or channels) that provide the most identity disentanglement, (3) manipulating these layers (or channels) in a desirable direction, and (4) generating a final image using the modified latent codes.}{steps.} 
We next define the key problems associated with each step.}{We next outline the four key steps of our approach for disentanglement in latent space.}
Details are provided in later subsections.

{\bf Steps 1+2 (Identity disentanglement):}
We first identify the channels/layers in latent space 
\revAA{that provides}{providing} 
the most desirable identity disentanglement. We call this the {\em identity disentanglement problem}. 

{{\bf Problem definition (Identity disentanglement):}
{\em \revBB{Given an input face and a GAN model with latent space $W$}{Given an input face $S$ and a GAN model $G$ with latent space $W$, assuming 
\revAA{}{that}
we have a projector 
\revAA{$\mathcal{P}: S^{W,H,D} \to W$, where ($W$, $H$, $D$) are the image's (width, height, depth)}{$\mathcal{P}: S \to W$} 
to project face $S \to L_S$}, the problem consists of 
\revAA{two sub-problems:
\begin{itemize}
    \item[1)] \revBB{Projecting the facial image onto the latent space $W$ in such a way that the reconstructed face has the smallest identity distance to the original image.}{Forming an identity disentanglement metric that measures both the change in identity and attribute of $S$ when manipulating $L_S$.}
    \item[2)] \revBB{Locating the channels in the latent codes (from step 1) that 
    provide the most identity disentanglement.}{Locating the layers/channels in the latent codes $L_S$ based on the metric (from step 1) that
    provide the most identity disentanglement.}
\end{itemize}}{identifying the layers/channels in the latent codes $L_S$ that
    provide the most identity disentanglement according to our disentanglement metric (equation (\ref{eqn:score})) that measures both the change in the identity and the attributes of $S$ when manipulating $L_S$.}}

\revBB{The first problem can easily be solved using pSp~\cite{richardson2021encoding}.  Here, we focus on the second problem.}{} 
To solve this problem, the main idea is to find 
\revAA{a}{an} 
identify direction that turns the identity of a face image away in such a way that other facial attributes are not affected.  Ideally, when the identified layers (or channels) are changed, only the identity should change while other (non-identity related) attributes are preserved.

Note that prior works that manipulate one feature at a time in latent space~\cite{shen2020interfacegan, harkonen2020ganspace, wu2021stylespace, patashnik2021styleclip, abdal2020image2stylegan++, voynov2020unsupervised} 
are not applicable in our context. 
One reason is that those attributes are easy to classify into binary classes or can easily be mapped to a scale between 0 and 1.  In contrast, with identity disentanglement there is no visible scale onto which all face images easily can be mapped.
To identify the layers to manipulate, we instead use our identity disentanglement metric to evaluate the disentanglement achieved when anonymizing a large set of images as per steps 3+4 (but without knowing the best layers to select).  However, before we present how this is done, we first present a brief introduction to how steps 3+4 are done when the best layers (or channels) already have been identified in step 2. 

{\bf Steps 3+4 (Feature-preserving anonymization):}
We next 
manipulate the identified layers (or channels) of the latent
\revAA{codes}{code $L_S$}
away from the original identity or towards an alternative identity 
\revAA{}{$T$}
that 
\revAA{have}{has} 
been randomly generated.
Finally, we use an encoder $G$ to generate an output image 
\revAA{}{$O$}
using the modified latent 
\revAA{code.}{code $L_{S'}$.}
We next formalize 
\revAA{the image anonymization}{the} 
problem these two steps aim to optimize.

{\bf Problem definition 
(Image anonymization
using latent
\hspace{-4pt}\break
code):} 
{\em Given a set of source 
\revAA{images,}{images $\mathbb{S}$,} 
the problem is to anonymize the images so that each of the anonymized images 
\revAA{}{$O_i \in \mathbb{O}$}
have maximum identity disentanglement distance from their respective source 
\revAA{images}{image $S_i$} 
given some desirable level of anonymity.}

Practically, this means that we would like to push the identity distance (weighted by $\alpha$ in equation (\ref{eqn:score})) as far away from the original identity while keeping the attributes (weighted by $\beta$) as close to the original identity as possible, given a desirable level of anonymity.  Here, the selected layers, the selected target 
\revAA{face,}{face $T$ (randomly generated),} 
as well as the distance that each code is pushed all impact the level of anonymity (and identity disentanglement) achieved.
\revBB{To evaluate the highest level of anonymity, we also consider the level of t-closeness and l-diversity achieved.}{}
\revAA{}{Fig.~\ref{fig:id-idsen} shows a high-level overview of the disentanglement steps taken to modify the latent code $L_S$ of an image $S$ using a subset of layers/channels of the latent code $L_T$ of a target identity $T$, and the evaluation of a resulting output image $O$.}

\commentout{
\begin{figure}[t]
\centering
\includegraphics[trim=0mm 0mm 0mm 2mm,  width=0.80\linewidth]{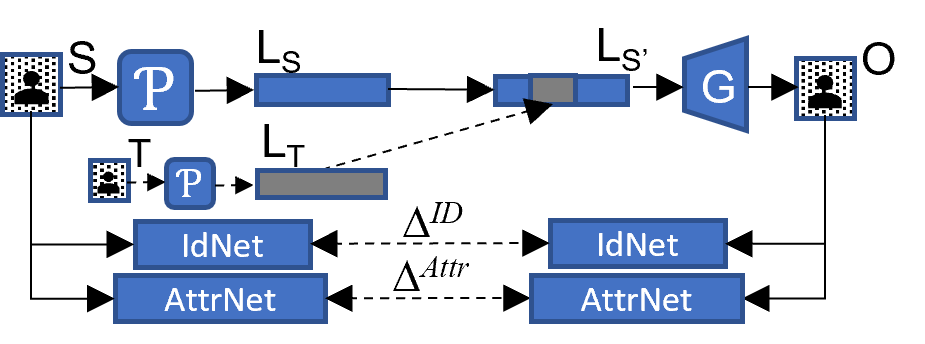}
\vspace{-12pt}
\caption{\revAA{Identity disentanglement in latent space}{Identity disentanglement approach in latent space.}}
\label{fig:id-idsen}
\vspace{-14pt}
\end{figure}
}

\subsection{Mapping to latent space (step 1)}

For the work presented here, we use a variant of StyleGAN called pSp~\cite{richardson2021encoding} that uses an enhanced latent space $W+$. 
One significant difference of pSp is that it has the capability to encode a real image to latent space.  After such a mapping has been done, the disentangled latent code can easily be used for image editing of different kinds.
In pSp, the layers of $\revBB{w}{W}$ codes are passed through a number of small fully connected convolutional networks to achieve the extended latent code in $\revBB{w+}{W+}$, which is proved to encode more information than in $\revBB{w}{W}$.
To simplify the notation, throughout the reminder of paper, we use the term latent space $W$ to refer to both $W$ and $W+$, and unless explicitly stated use the latent codes provided by pSp.

\begin{figure*}[t]
\begin{minipage}[t]{0.32\textwidth}
    \centering
    \vspace{-36pt}
\includegraphics[trim=0mm 0mm 7mm 2mm,  width=0.98\linewidth]{newFigs/pets23a-figure2b-v03.png}
\vspace{-8pt}
\caption{\revAA{Identity disentanglement in latent space}{Identity disentanglement approach in latent space.}}
\label{fig:id-idsen}
\vspace{-14pt}
\end{minipage}
\hfill
\begin{minipage}[t]{0.67\textwidth}
\centering
    \begin{subfigure}{0.48\textwidth}
        \raggedleft
        \includegraphics[width=0.82\linewidth]{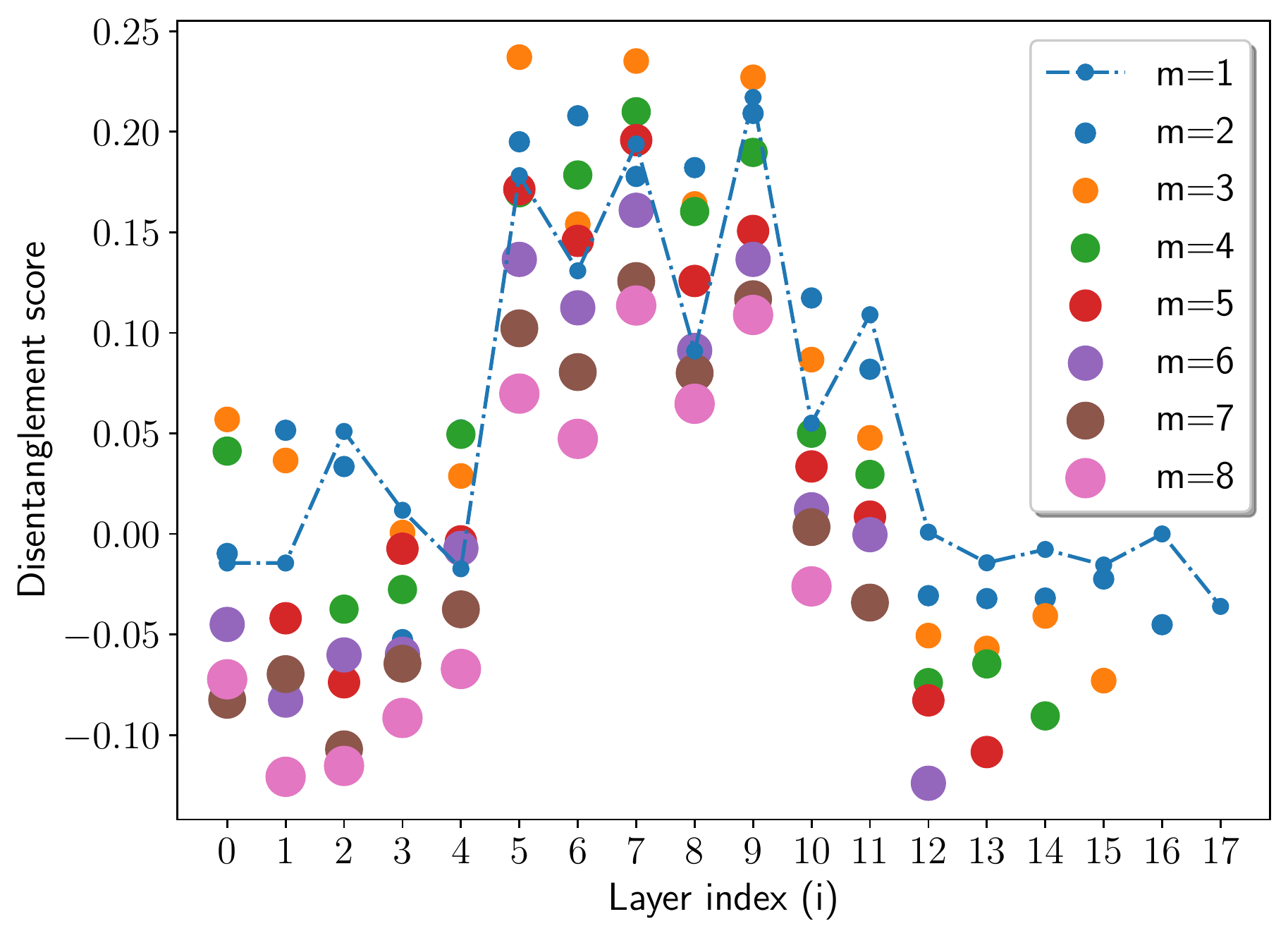}
        \vspace{-6pt}
        \caption{Scoring for different subsets of layers}
        \label{fig:score_breakdown}
    \end{subfigure}
    \hfill
    \begin{subfigure}{0.48\textwidth}
        \raggedright
        \includegraphics[trim = 0mm 16mm 0mm 0mm, width=0.99\linewidth]{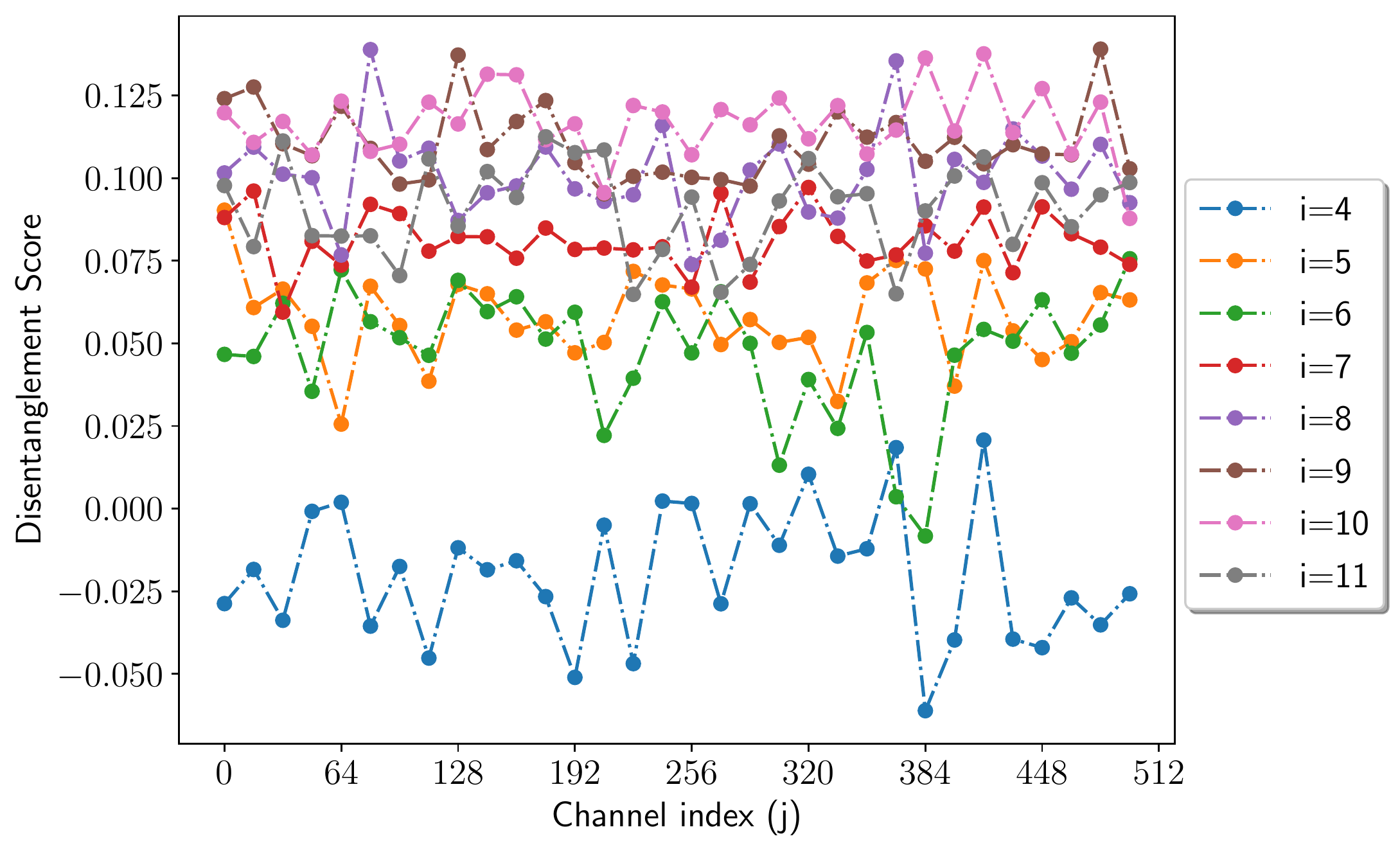}
        \vspace{-6pt}
        \caption{Scoring for each channel}
        \label{fig:score_channels}
    \end{subfigure}
    \vspace{-10pt}
    \caption{Disentanglement scores of swapping layers and channels}
    \vspace{-12pt}
\end{minipage}
\end{figure*}

\commentout{
\begin{figure*}[t]
    \begin{subfigure}{0.48\textwidth}
        \raggedleft
        \includegraphics[width=0.66\linewidth]{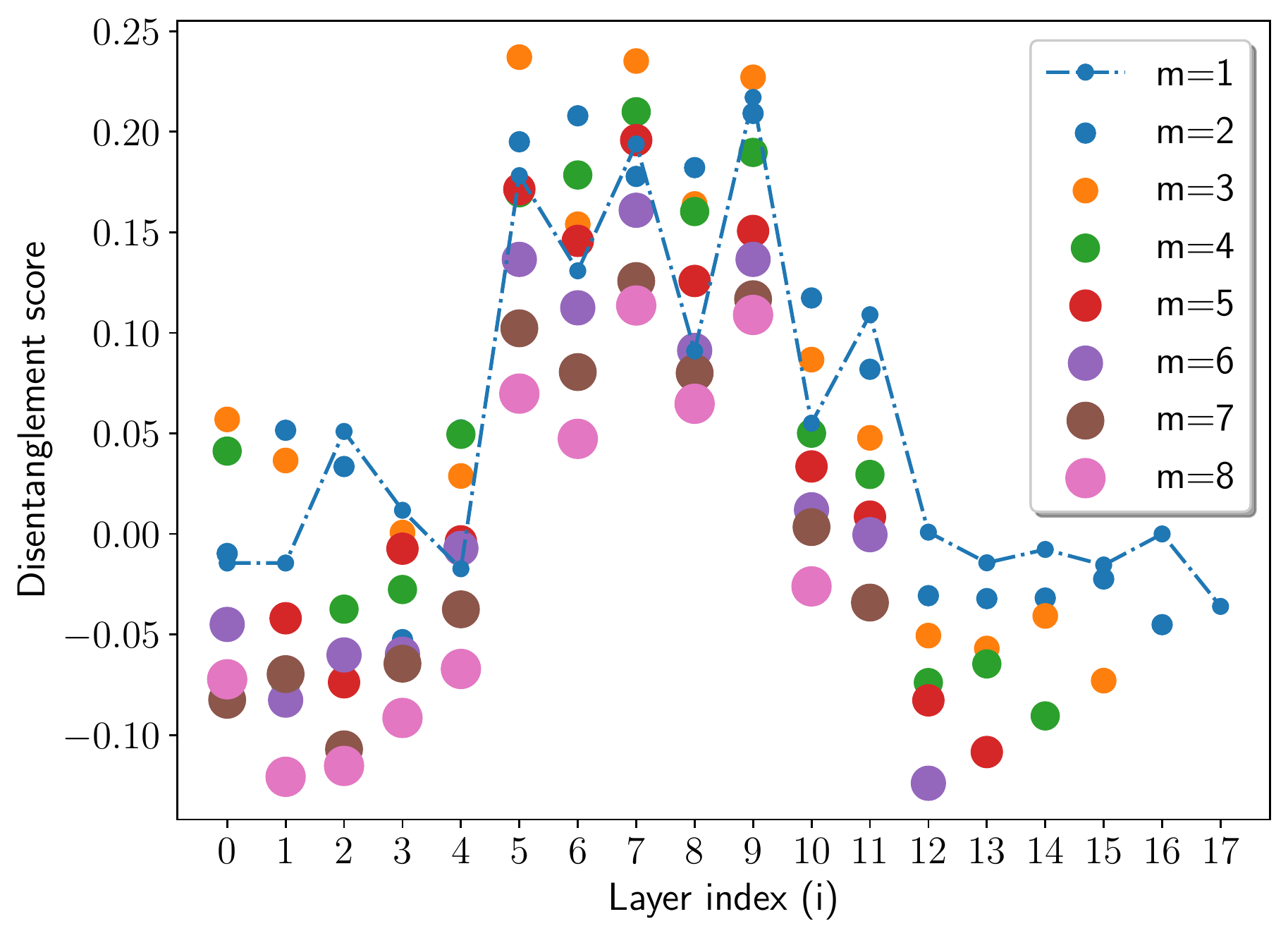}
        \vspace{-6pt}
        \caption{Identity disentanglement for different subsets of layers}
        \label{fig:score_breakdown}
    \end{subfigure}
    \hfill
    \begin{subfigure}{0.48\textwidth}
        \raggedright
        \includegraphics[width=0.79\linewidth]{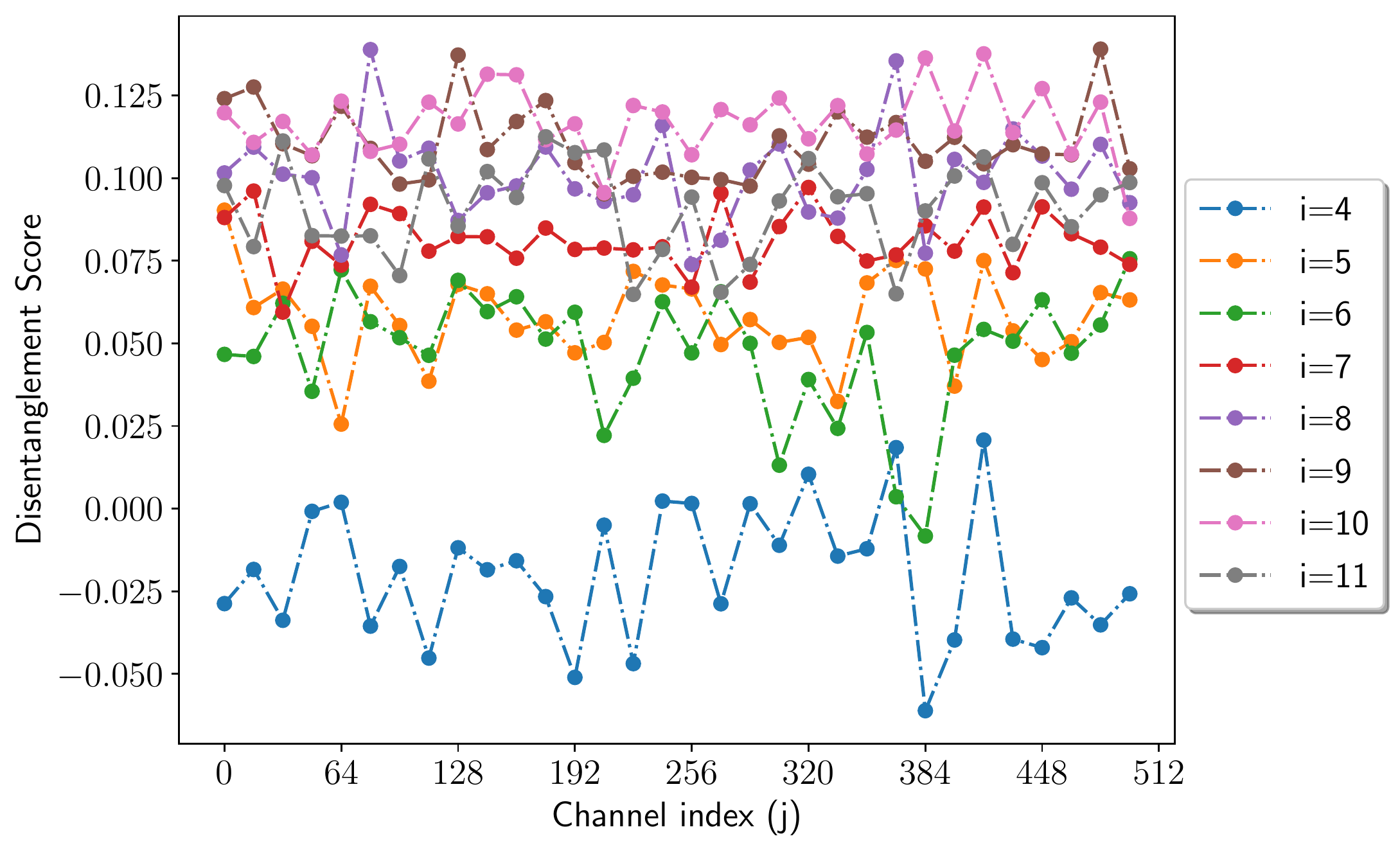}
        \vspace{-6pt}
        \caption{Scoring for each channels in latent codes}
        \label{fig:score_channels}
    \end{subfigure}
    \vspace{-10pt}
    \caption{Disentanglement scores of swapping layers and channels}
    \vspace{-12pt}
\end{figure*}
}

\subsection{Finding layers to modify (step 2)}

Our method for finding the best subset of layers (or channels) to modify in latent space is to evaluate the disentanglement that a large number of random pairs of faces $\revBB{A}{S}$ and $\revBB{B}{T}$ achieve when swapping different subsets of their respective latent codes $\revBB{w_A}{L_S}$ and $\revBB{w_B}{L_T}$.

The evaluation of one such test is illustrated in Fig~\ref{fig:id-idsen} and the results are presented in Sec.~\ref{sec:layers-eval}. 
\revBB{}{In the figure, $L_S$ and $L_T$ are the source and target latent codes corresponding to the source $S$ and target identity $T$ (for the purpose of anonymization the target identity is randomly generated by a face generator $G$). The gray area of the latent 
code $L_{S'}$ illustrates the swapped layers.
To estimate the identity disentanglement score we take the result $L_{S'}$ 
(equal to $L_S$ where the identity is swapped to the random identity in $L_T$),
and pass it through the face generator $G$. The corresponding generated faces from $L_S$ and $L_{S'}$ 
are
then be passed through $IdNet$ to calculate the identity distance 
and $AttrNet$ to calculate attribute distance. 
Finally, those distances are used to calculate the identity disentanglement scores.}
\revAA{}{By symmetry, 
to evaluate
the impact of swapping 
layers, we 
do the same swapping and evaluation of the impact this has on $T$. For clarity, this is omitted from the figure.}

To keep this analysis feasible (there is 
\revAA{an exponential number of possible combination}{exponentially many combinations} 
of layers and evaluation is time consuming), we 
\revAA{considered the use of}{used} 
a window-based approach in which we swap groups of $m$ consecutive layers; i.e., all layers from layer $i$ through layer $i+m-1$.  The use of consecutive layers is further motivated by having found that neighboring layers often contribute to similar (or the same) features.  We then select the identity direction as the $(i,m)$ values that maximizes the average identity disentanglement score (calculated using equation (\ref{eqn:score}) between a large set of 
\revAA{origin}{source} 
images $\revBB{A}{S}$ and the generated output images $\revBB{O=G(w_{A'})}{O=G(L_{S'})}$ obtained \revBB{if}{when} swapping layers 
$\mathcal{L}_{i,m} = (i, i+1, ..., i+m-1)$ 
of latent code $\revBB{w_A}{L_S}$ with the same layers of the latent code $\revBB{w_B}{L_T}$ of some random image $\revBB{B}{T}$.

\subsection{Manipulating the identity (step 3)}

We consider two ways of manipulating a selected set of layers 
$\mathcal{L}$
in the latent code:
    (1) by swapping layers 
    $\mathcal{L}$
    of the source identity $S$ with the same set of layers of a target identity $T$, and 
    (2)
    by pushing the source identity $S$ in some direction to obtain a new identity $S'$.  

For the first case, we use the same swapping methods as discussed in the previous subsection and illustrated in Fig.~\ref{fig:id-idsen} but this time with $S$ and $T$ as the two \revBB{original}{input} images.
For the second case, we simply pick the identity $S'$ that maximizes the identity disentanglement score from a large set of randomly generated images where we manipulate a selected set of layers in the latent code.  Here, we select to modify the set of layers that we found provided the best results 
\revAA{when performing swapping for}{over} 
a large number of example images.  

\revCut{We next describe the dataset we used for our analysis and how we identified the layers that provide the best identity disentanglement.} 

\subsection{\revAA{Dataset, pre-training, and our training-evaluation split}{Dataset and training-evaluation split}}

Unless explicitly stated, we use the \revBB{CelebA~\cite{liu2015faceattributes}}{CelebAMask-HQ dataset~\cite{lee2020maskgan}} dataset in our experiments. The main reason for this choice is that each face image in the dataset is annotated with 
\revBB{both identity and 40 binary attributes}{identity, 40 binary attributes, 
and it comes with segmentation masks that we use for our analysis in Sec.~\ref{sec:mask}}. In total, the dataset includes 
\revBB{202,599}{30,000} 
images of 
\revBB{10,777}{6,217} 
identities.
\revCut{This makes it one of the facial image datasets with most identities.}
In addition, the annotated attribute values provide ground truth for the training of our $AttrNet$ model.
Finally, we have found that the dataset's high resolution is valuable for ensuring low identity loss when projecting a face onto latent space $W$.

For our evaluation, we split the dataset into three non-overlapping sets of identities: the source set, the target set, and a validation set 
\revAA{that we only used for validating our results.}{(used only for validation).}  
The first two sets (source + target) each include 
\revAA{4,000}{2,000} 
unique  
\revCC{identities, while the}{identities ($\approx$10,000 images).  The}
validation set contains the remaining 
\revBB{2,777}{2,217} 
identities.  For simplicity, 
\revCC{we represent each identity using (only) a random sample image for that individual.}{each identity is represented using (only) a single random sample image.}

\revCut{
\begin{figure}[t]
\centering
\includegraphics[width=0.8\linewidth]{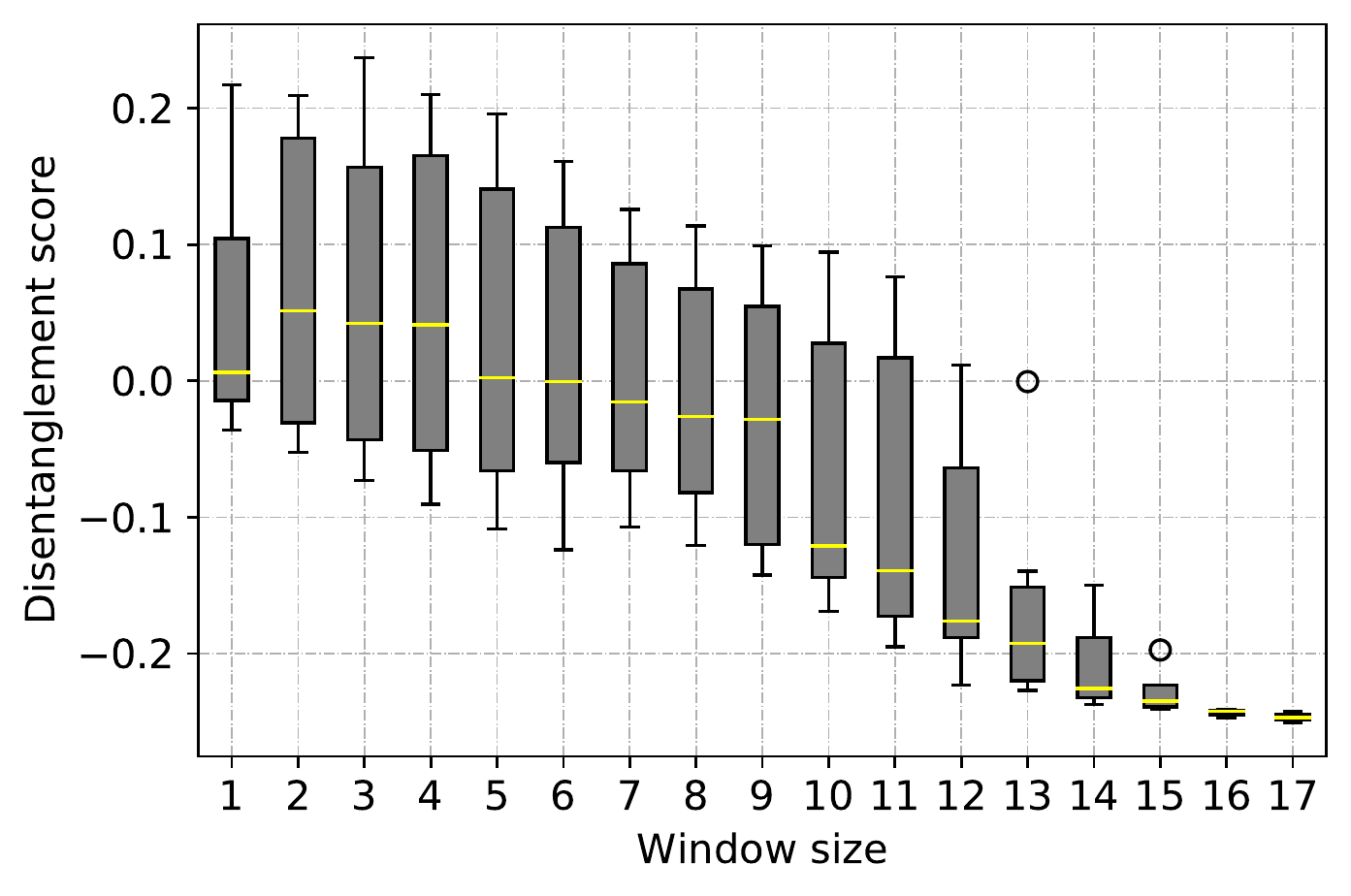}
\vspace{-8pt}
\caption{Disentanglement scores for different window sizes $m$}
\label{fig:score_layer}
\vspace{-6pt}
\end{figure}}

\subsection{Results when swapping layers}\label{sec:layers-eval}

We have found that modifying 2-to-4 layers often produce the best results.  
\revCut{This is exemplified by Fig.~\ref{fig:score_layer}, where we show the median disentanglement scores (yellow lines) with different window sizes $m$. Here, we have aggregated the results over all possible starting points $i$. In addition to the median values, the boxplot also shows the maximum and minimum scores (whiskers) and the 75\% and 25\% percentiles (top and bottom of the boxes).}
\revAA{Looking closer at}{Comparing} 
the individual starting points $i$, we have found that layers 5-9 typically provide the best results, with layers 5, 7, and 9 (individually) contributing the most to the disentanglement.  This is shown in Fig.~\ref{fig:score_breakdown}, where we show the disentanglement scores for different combinations of starting points $i$ (on x-axis) and window sizes $m$ (markers of different color and size).  We use a dotted line to show the baseline when only switching a single layer 
\revAA{(i.e., $m=1$).}{($m$$=$$1$).} 
From the figure it is clear 
that the best choice (on average) is to use layers 5-7 (assuming we use consecutive layers).  Furthermore, the distinct spikes seen for layers 5, 7 and 9 show that these layers individually provide the highest identity disentanglement.  

\revAA{The later}{The} 
last observation 
\revAA{above raises}{raises} 
the question whether greedily using layers (5,7,9), would improve the results over using the best consecutive layers (5,6,7).  To answer this question, 
\revAA{we have run experiments with this greedy configuration.  
\revAA{In our evaluation, we therefore}{We next} 
compare 
\revBB{}{visual results of} 
these two candidates head-to-head.}{we have compared the two  candidates head-to-head and present visual results next.}

\begin{figure}[t]
\centering
\rotatebox{90}{\parbox{6.2cm}{\phantom{-----}{\small(5,7,9)} \phantom{----------} {\small (5,6,7)} \phantom{--------} {\small Target} \phantom{---------}{\small Source} }}
\includegraphics[width=0.95\linewidth]{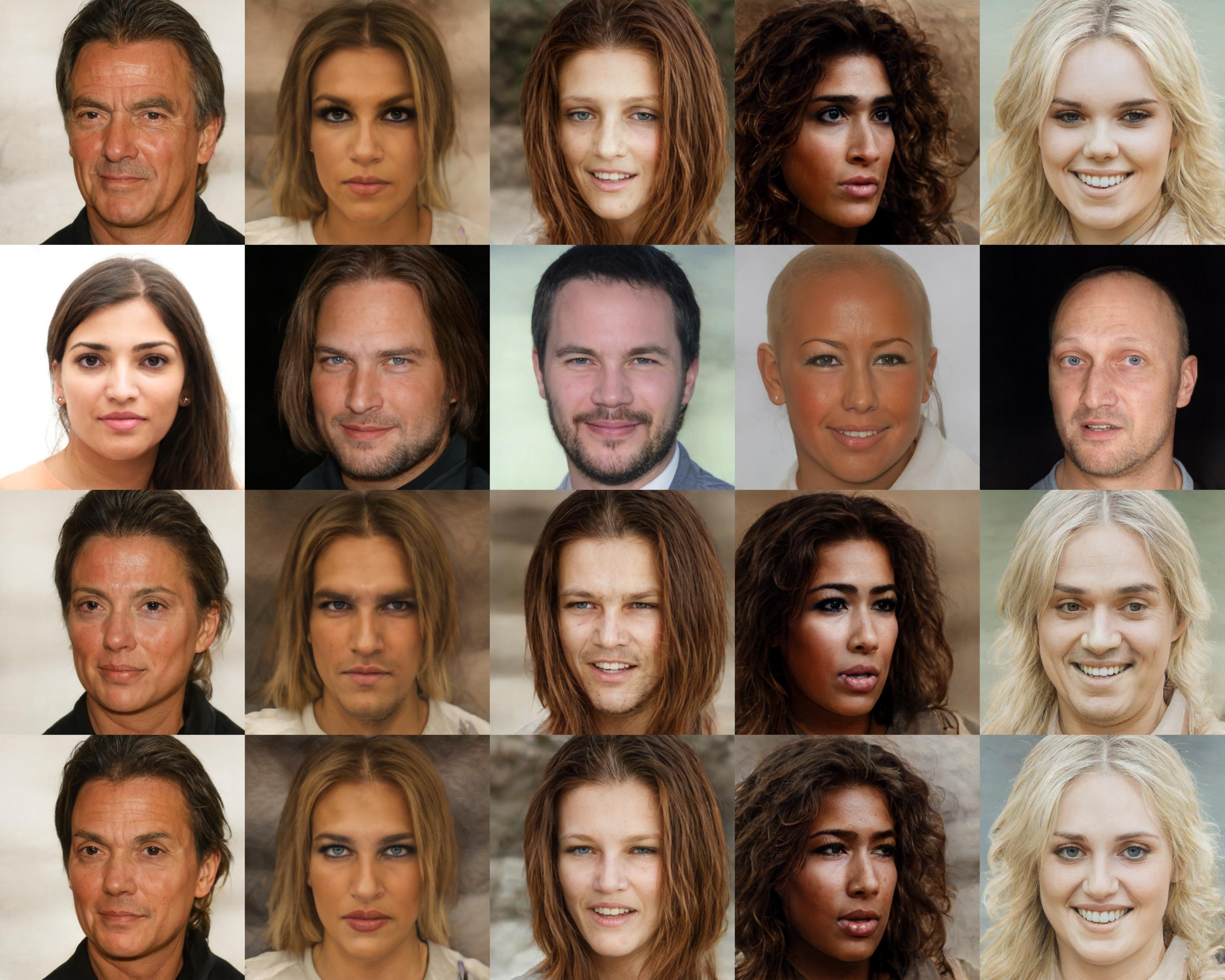}
\vspace{-8pt}
\caption{Example 
\revAA{anonymization by}{results} 
swapping different layers of the latent codes of the 
\revAA{source identity}{source} 
(top row) and 
\revAA{target identity}{target} 
(second row).
\revAA{The results}{Results} 
swapping layers (5,6,7) are shown on the third row and 
\revAA{the results}{results} 
swapping layers (5,7,9) are shown on row four.}
\label{fig:swap-layers}
\vspace{-12pt}
\end{figure}

\subsection{Example results}\label{sec:picking_layers}

Fig.~\ref{fig:swap-layers} shows representative example results (bottom two rows) generated using the latent code $\revBB{w_G}{L_O}$ obtained by greedily replacing 
\revAA{the three}{the} 
top-three individual layers (5,7,9) or the top-three consecutive layers (5,6,7) of the latent code $\revBB{w_{S}}{L_S}$ of the source face (top row) with the corresponding layers of the latent code $\revBB{w_{T}}{L_T}$ of the target face (second row). 
\revAA{The faces shown here corresponds to representative examples of random faces selected}{The example faces were randomly selected}
from the source and target 
\revAA{sets of our dataset.}{sets.}
\revAA{As desired, when using consecutive layers (third row), most non-identity related attributes (e.g., hair, face expression, pose, eye direction, skin tone, etc.) are preserved in all the samples, while some subtle facial features are moved towards the identity of the target face. Perhaps most importantly, these changes are sufficient to prevent facial recognition tools from identifying the original source identity.}{While the approach of swapping layers allows us to move the identity towards that of the target, we have observed some weaknesses.} 
\revAA{One problem}{First,} 
we have found 
\revAA{with this approach is that the}{that the attributes such as the age and gender often are dependent on the target face.  Since this is generated randomly, a naive implementation may in some cases therefore change gender and age.} 
\revBB{}{We provide further analysis on the correlation of identity and individual attributes in Appendix A.}
\revAA{}{To address this shortcoming, we provide two solutions: (1) a greedy approach to optimally preserve all attribute (described in the next section) and (2) the use of available semantic editing frameworks to control the attribute (discussed 
\revDD{further in}{in} 
Sec.~\ref{sec:preserving_attr}).}
Second, 
\revDD{the 
background sometimes changes. For example, while}{while} 
the background is 
\revDD{\revBB{}{relatively} 
identical}{similar} 
for 
\revAA{most of the}{most} 
cases, there 
are clear 
\revDD{\revBB{exceptions}{changes} 
\revBB{(e.g., sixth column)}{in the second column}.}{changes (e.g., in second column).}
To handle these cases,
we suggest combining the results with pixel-space 
\revShort{manipulations.  Segmentation masks such as those that we use in Section~\ref{sec:mask} can be used for such purposes.}{manipulations; e.g., as with the segmentation masks used in Sec.~\ref{sec:mask}.}

\revDD{Finally, we note that the results shown in the bottom row (which greedily switches the best individual layers) better preserve the properties of the source images whereas the results in the results in third row (i.e., consecutive layers) shows a bigger visual push towards the target identity.  In general, we have found that greedily switching layers typically provide more control of the attributes being preserved. Having said that, in both cases, the identity has been moved closer to the target identity, while preserving the majority of facial features of the source image.}{Finally, while both approaches preserve the majority of facial features of the source image, we have found that greedily switching the best individual layers (bottom row) better preserves the properties of the source images and that switching consecutive layers (third row) results in a face somewhat closer to the target identity.} 

\section{Channel manipulation}\label{sec:channels}

In the context of non-identity related features, prior 
\revAA{researches have}{research~\cite{wu2021stylespace, harkonen2020ganspace} has}
found that more fine-grained control of facial features 
\revAA{are}{is} 
possible when manipulating individual channels.  In this section we investigate to what degree it may (or may not) be beneficial to swap a subset of channels rather than the full layers.

 \cutAA{
 \subsection{Pixel-space-aware disentanglement}

To achieve sufficient anonymization, we have found that it is important to manipulate a significant number of channels.  Since far from all relationships between individual channels are known, it is perhaps not surprising that some changes may result in undesirable effects in the pixel space. We have therefore found that it is important to take into account the visual appearance also in the pixel space when selecting a set of channels to manipulate. 
For the purpose of selecting which channels to swap (or manipulate), we have therefore incorporated the use of the Frechet Inception Distance (FID)~\cite{karras2019style} during the channel (and identity) selection process.

{\bf High-level greedy approach:}
Consider first the case when we swap channels of two identities $S$ and $T$. 
In this case we, pick the set of channels that maximizes the equation: 
\begin{align} \label{eqn:pixel-score}
    IP_{score}(S,O) & = \alpha \cdot h(\delta(IdNet(S), IdNet(O)) \nonumber\\
    & ~~~ - \beta \cdot h(\delta(FID(S,O)),
\end{align}
where $O$ corresponds to any output image generated after swapping some subset of channels between image $S$ and the target identity $T$.  To find a good set of channels to swap we applied different greedy searches at different block granularities. 

While equation (\ref{eqn:pixel-score}) was used for the selection process, it is important to note that we still used our (original) identity disentanglement scoring metric, defined in equation (\ref{eqn:score}), for the evaluation of the final selection.
}

\subsection{\revBB{Individual}{Swapping individual} channels}

Let's first consider the impact that each channel has on the identity disentanglement score.
Fig.~\ref{fig:score_channels} shows the results broken down for each of the (intermediate) layers that we found provided the 
\revAA{best}{most} 
disentanglement.  To ease visualization, we 
\revAA{applied a moving average function with a window size of 16}{show the average over blocks of 16 channels}.
\revAA{}{When comparing 
Figs.~\ref{fig:score_channels} and~\ref{fig:score_breakdown}, we make several interesting observations.
First, swapping channels appears to provide more effective identity disentanglement than swapping layers.  For example, while the observed disentanglement score when swapping\revCC{16}{} channels (e.g., peak of 0.125 in Fig.~\ref{fig:score_channels}) is smaller than the observed disentanglement score when swapping 1-to-8 layers (e.g., 0.25 in Fig.~\ref{fig:score_breakdown}), the rate of change in the disentanglement score per channel is higher when considering that a single layer includes 512 channels.} 
\revAA{In general, we observe significant differences between the disentanglement achieved by swapping different channels and several of the top spikes are associated with some of the layers that achieved the largest identity disentanglement (if switched as a whole).}{
Second, we observe significant differences between the disentanglement achieved by swapping different channels.  Third, several of the top spikes (Fig.~\ref{fig:score_channels}) are associated with some of the layers (Fig.~\ref{fig:score_breakdown}) that achieved the largest identity disentanglement (if switched as a whole).  However, we also see some clear differences.  For example, the best scores when considering consecutive layers (Fig.~\ref{fig:score_breakdown}) belongs to layers (5, 6, 7).  However, swapping\revCC{16 consecutive}{} channels in these layers typically results in smaller disentanglement than swapping\revCC{16 consecutive}{} channels for the higher numbered layers 8, 9 and 10 
(Fig.~\ref{fig:score_channels}). These observations also show that the disentanglement score is not additive.}

\begin{figure}[t]
\centering
\includegraphics[width=0.68\linewidth]{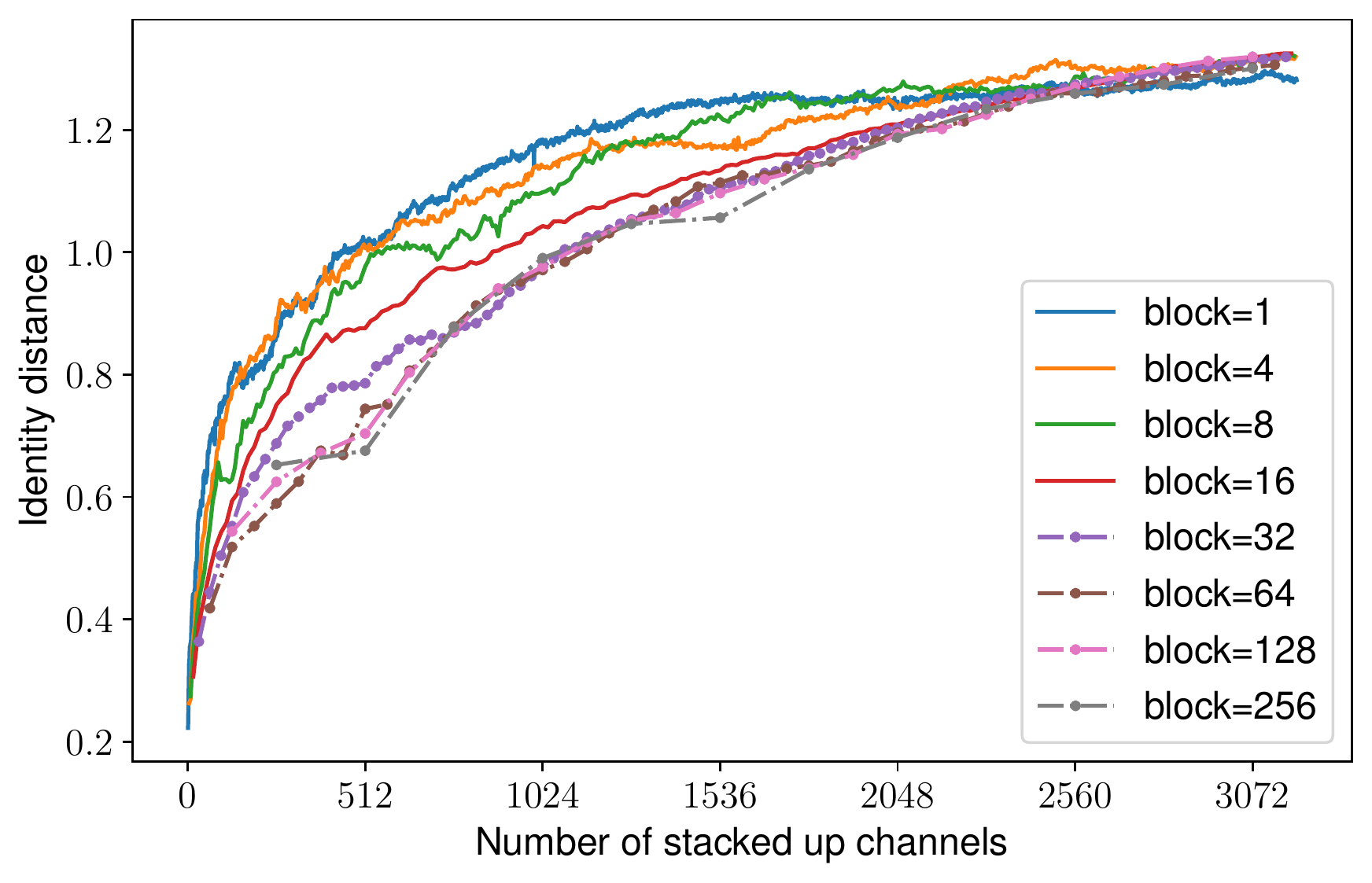}
\vspace{-12pt}
\caption{Block size on identity distance}
\label{fig:block_vs_id}
\vspace{-12pt}
\end{figure}

\subsection{Block-size evaluation}

We next consider the number of channels that must be switched to achieve different levels of anonymity and the impact of grouping individual channels into blocks that either are switched or not. Fig.~\ref{fig:block_vs_id} summarizes these results.  Here, we show the identity distance between the source face and the generated output face.

A few observations are noteworthy.
First, there are diminishing returns with the gains of most block sizes flattening out after greedily swapping approximately 2,000 channels (on a per-block basis). Here, it should be noted that this corresponds to roughly four layers (which together have 
\revAA{2,056}{2,048} 
= 4 $\times$ 512 channels).
Second, the difference between using block sizes of 32-256 is small (and similar to switching entire layers, which contain 512 channels).  This suggests that we either may want to use blocks smaller than 32 or we might as well swap entire layers.
Third, while the smaller block sizes see the fastest improvements in identity distance, there is an infliction point around 2,600 channels (similar to 5 layers) 
\revDD{were}{where} 
larger block sizes are able to achieve larger identity distance.  At this point the identity distance is around 1.25 \revBB{}{(a threshold at 99\% accuracy of facial recognition model ArcFace~\cite{deng2018arcface} on LFW benchmark \cite{huang2008labeled})}.  Since the larger block sizes often result in visually more appealing images, for cases where we want even greater anonymity than 1.25, it is therefore typically desirable to swap layers rather than individual channels.  On the other hand, if the required level of anonymity is less, then less channels are needed to be switched if using smaller block sizes. 
Finally, for the case when using the anonymity threshold of 0.9 (%
a threshold at 95\% accuracy of facial recognition model ArcFace~\cite{deng2018arcface}), anonymization is easily achieved with any block size, although significantly fewer swaps are needed to achieve this threshold when using smaller block sizes.

\begin{figure}[t]
\centering
\rotatebox{90}{\parbox{4cm}{\phantom{-----}{\small Result} \phantom{-----} {\small Target} \phantom{-----} {\small Source} }}
\includegraphics[width=0.9\linewidth]{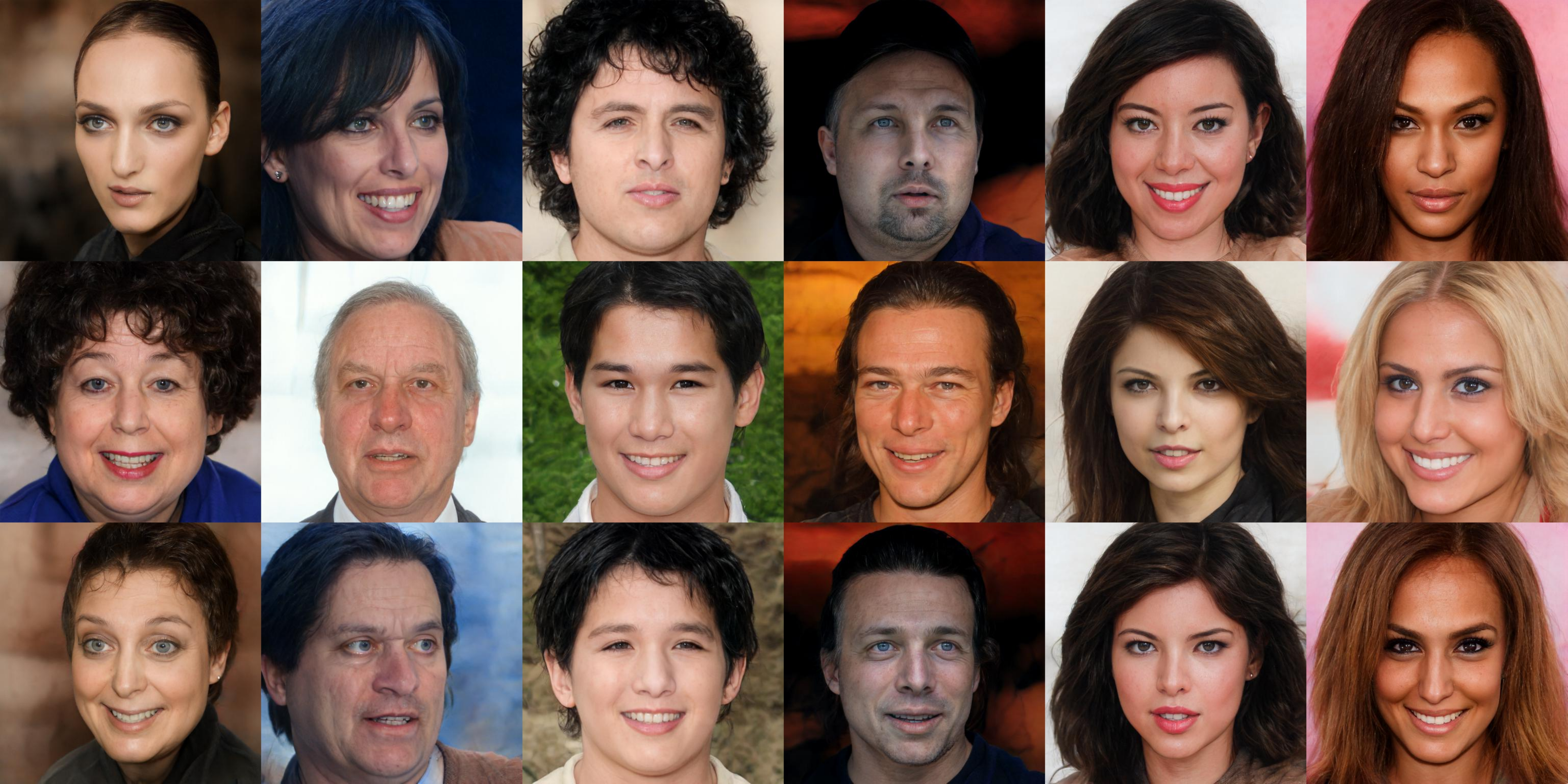}
\vspace{-8pt}
\caption{Example result using coarse-grained channel swapping.  Here, we swap the 8 top-scoring blocks 
\revAA{with size of}{of size} 
256. The three rows correspond to the 
\revAA{source image}{source} 
(top row), 
\revAA{target image}{target} 
(middle row), and 
\revAA{}{the}
generated output image (bottom row).}
\label{fig:swap_latent_channels}
\vspace{-14pt}
\end{figure}

\subsection{Visual example results}

We now turn to the visual results. 
Consider first (relatively) course-grained channel swapping.
Fig.~\ref{fig:swap_latent_channels} shows example results when swapping the 8 top-scoring blocks with size of 256.  As desired, our method is able to generate an output identity (bottom row) 
\revAA{that substantially have moved the identity of the source identity (top row) towards that of the target identity, while preserving many of the facial properties of the source images.}{that has been pushed away from the source identity (top row) and towards the 
\revAA{}{(random)}
target identity (second row), while preserving many of the facial properties of the source image.}
\revCC{However, compared to changing entire layers we have found that we have less control of what features are preserved using this greedy approach.  For example, while the pose and general facial shape typically are preserved, several other features become a blend of the two faces as the identity is being pushed towards that of the target identity. Examples here include the hair and facial expression. We also found that the exact features being preserved and those that may become a blend vary more from case-to-case than when switching layers.}{}

While using more channels or more \revBB{fin-grained}{fine-grained} 
channel swapping 
\revAA{clearly result}{result} 
in less control of what features 
\revAA{being}{are} 
preserved, we have found that such channel swapping still can be effective in hiding the identity from facial recognition tools.  For example, consider 
\revAA{next the}{the} 
case when we treat each channel independently (i.e., block size of 1).
\revCut{As described in prior sections, we greedily picked the highest scoring candidate identity $S'$ from a set of candidate identities obtained by swapping the top channels of the source image with randomly generated faces.}  
Fig.~\ref{fig:top_1000} shows example results where we swapped the 
\revBB{1,000}{500-2,500} 
top channels. 
\revAA{Here, the identity does}{Notably, with less than 1,000 channels swapped, both the identity and attributes do} 
not change significantly to the human 
\revAA{eyes,}{eye,} 
but change enough to trick facial recognition tools (ensured by the optimization done using $IdNet$).
For example, the identity distance 
\revBB{}{for 1,000 channels} 
in all cases exceeds the minimum threshold of 
\revBB{0.9 (which is the threshold that gives us 95\% accuracy using an ArcFace model on the LFW dataset)}{0.9} 
\revCC{}{(see Fig.~\ref{fig:block_vs_id}).} 
\revCC{In particular, the scores obtained for the 
\revBB{six}{four} 
pairs above were: 1.04, 0.90, 0.98, 
\revAA{0.90, 0.95, and 1.00.}{and 1.00.}
This}{This}
demonstrates that the solution approach can be useful for somebody that wants to post images that allow their friends to recognize them but that still trick facial detection tools. \revBB{}{When we swap more than 1,000 channels, %
increasingly many
identity features and attributes 
are taken from the target images. However, important attributes such as gender, facial expressions, pose, lighting are well preserved. Even in the case when we swap 2,500 channels, we see that context features such as background, hairstyles, lighting are well preserved. Compared to swapping layers, this case could still be considered as an improvement.}

On the 
\revAA{more negative}{negative} 
side, 
\revAA{with the channel-based approach, we see some variations in the way that each image was anonymized (demonstrating loss of control of which features are preserved when using fine grained blocks). While for some of the images almost 
all the attributes are the same, in a few cases we observe noticeable changes in facial expression.}{we lose some control of which features are preserved when using fine-grained blocks. This is observed by somewhat more noticeable variations in the way that each image is anonymized.  For example, while almost all attributes are the same for some images, there are a few cases with noticeable changes in the facial expression.} 
However, in general, the method 
\revAA{is able to maintain}{creates} 
relatively similar images while still fooling facial detection techniques. 

\begin{figure}[t]
\vspace{-6pt}
    \centering
    \rotatebox{90}{\parbox{10cm}{
                                \phantom{--------}{\small 2500}
                                \phantom{---------}{\small 2000}
                                \phantom{---------}{\small 1500}
                                \phantom{---------}{\small 1000}
                                \phantom{---------}{\small 500}
                                \phantom{---------} {\small Target}
                                \phantom{--------} {\small Source} }}
    \includegraphics[width=0.96\linewidth]{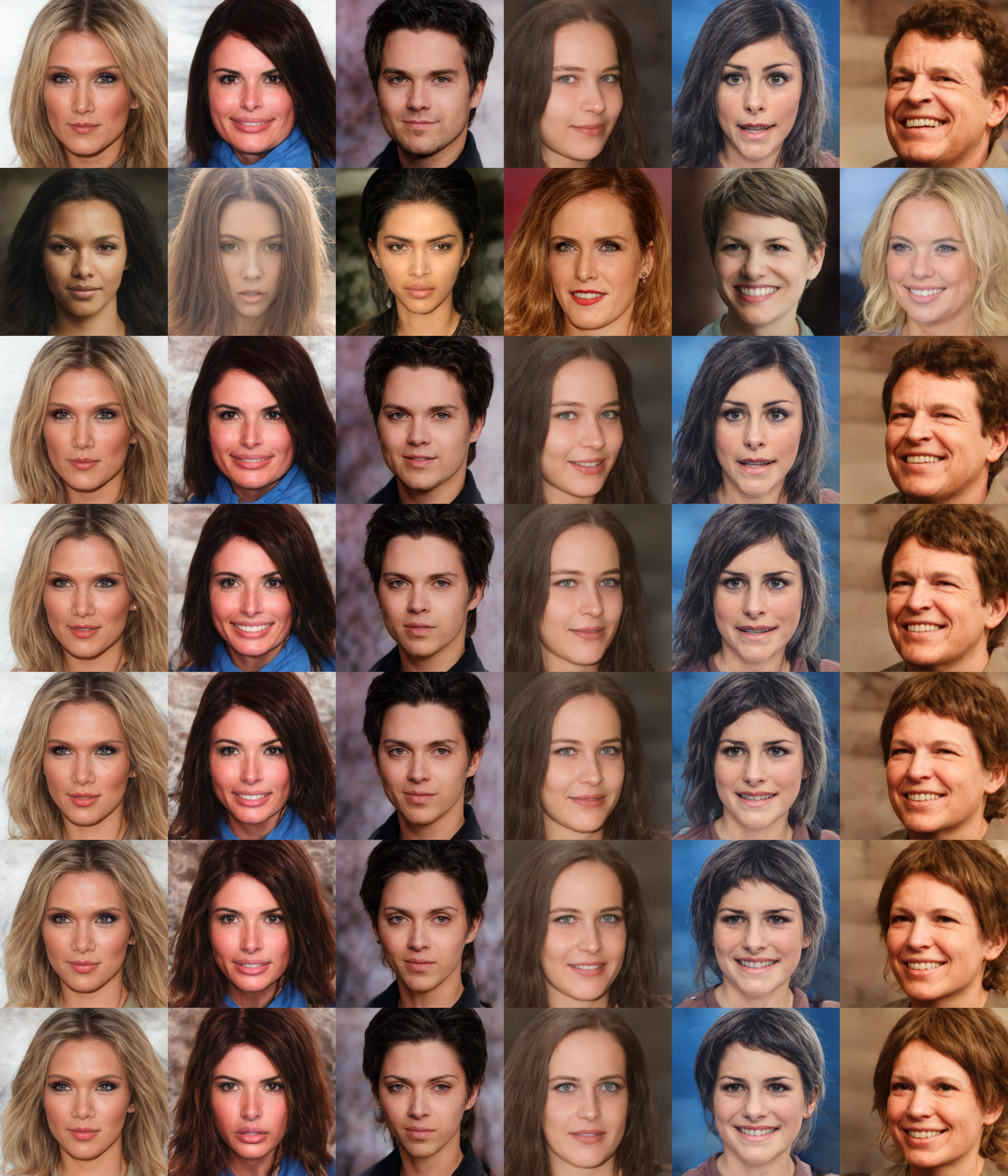}
    \vspace{-18pt}
    \caption{Example result using fine-grained greedy channel swapping.
    Here, we swapped the 
    \revAA{1,000 top channels of the source image with a set of 4 random identities.  Source image (bottom row) and the anonymized identity (bottom row) are shown.}{500-2,500 top channels of the source (top row) with a random target (second row).}}
    \label{fig:top_1000}
    \vspace{-14pt}
\end{figure}

\section{Feature-aware identity masking}\label{sec:mask}

While the 
\revDD{anonymization solutions}{solutions} 
presented in previous sections provide 
\revDD{a very}{an} 
effective method for 
\revDD{anonymization of}{anonymizing} 
individual images, these solutions are not sufficient on 
\revDD{their own if wanting to preserve the background and may result in odd effects if applied on a video.  To handle these contexts,}{their own. To better preserve the background and to address some odd effects if applied in the video context,} 
we propose a method that incorporates the use of segmentation masks.

{\bf High-level approach:}
Our key idea is to generate a face that has the same pose as the source image and then swap a selected part of this face with the randomly generated target face. To do so, we first extract the face segmentation mask of the source image, use this face mask to generate a target face that has a similar segmentation mask, and then swap a selected part of the source face and the target face. Fig.~\ref{fig:mask} presents an overview of this approach.

\begin{figure}[t]
\centering
\includegraphics[width=0.96\linewidth]{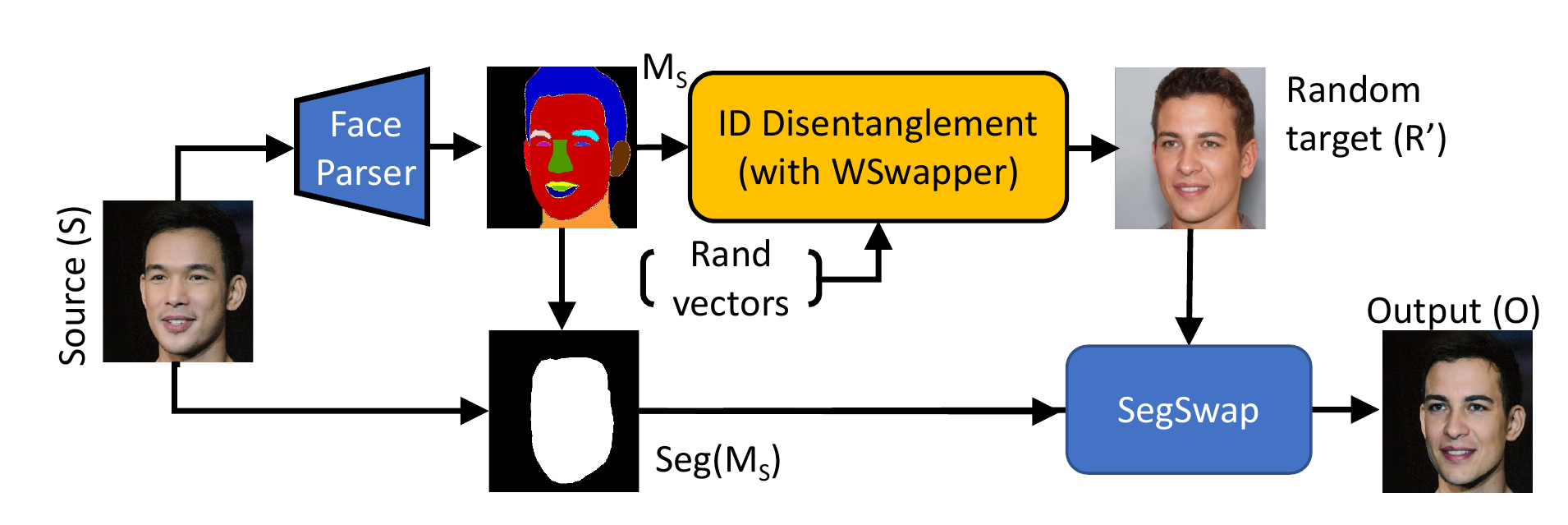}
\vspace{-12pt}
\caption{Feature-aware identity \revAA{masking with segmentation mask}{masking}}
\label{fig:mask}
\vspace{-12pt}
\end{figure}

{\bf Mask-based swapping:}
This is a non-trivial task but one for which fortunately several face-swapping models already provide excellent support. In fact, this problem can 
\revAA{}{in part}
be well handled by face-swap models such as faceswap~\cite{deepfake}, DeepFaceLab~\cite{perov2020deepfacelab}, FSGAN~\cite{nirkin2019fsgan}.  The big difference between the existing face-swap models and our framework is that while their objective typically is to blend in a target identity, we try to hide the identity of the source face. In particular, face-swap solutions try to turn the face into that of a specific target identity, while we have less strict requirements on the final identity's relationship to the target identity.  
\revAA{However, instead we have much higher need for the results to hide}{Instead we have a higher need for hiding} 
the source identity, while still preserving facial features and producing images of high utility. 

One important question that arise here is how much of the facial area needs to be changed to allow different degrees of 
\revAA{anonymity (i.e., to either trick both humans and facial recognition tools or only the later).}{anonymity.} 
Fortunately, as seen in 
\revAA{the previous section,}{Secs.~\ref{sec:layers} and~\ref{sec:channels},} 
most of the disentanglement were achieved by
manipulating latent codes that preserved the pose and mostly affected central areas around the eyes, nose, and mouth. These areas are often well captured by segmentation masks.

{\bf Mask generation:}
For the generation of a precise segmentation mask, we use existing 
GAN segmentation models, including MaskGAN~\cite{lee2020maskgan} and pix2pixHD~\cite{wang2018high}.  These models are becoming widely used and have been shown to achieve high accuracy and performance. The use of existing models also avoids the need to train new GAN models (which can be highly time consuming). 

{\bf Generation of random face:}
\revAA{Ensuring that the two faces have a shared segmentation mask, significantly simplifies swapping identity and blending in the (new) random face. Given}{Given} 
a segmentation mask, the next important step is 
\revAA{therefore to}{to} 
generate a random face that shares the same segmentation mask. One straightforward approach to do this is to use the segmentation mask as input to StyleGAN's encoder (e.g., pSp~\cite{richardson2021encoding} in our case) in a similar way as with in-painting methods~\cite{maximov2020ciagan, wu2019privacy}.
However, this approach does not work well for our purposes since a given segmentation mask often is based on a particular face. If that face was used to train the model, the generated face is therefore likely to closely resemble the original identity. 
In contrast, for
the purpose of anonymization, 
we need the newly generated face
to be completely random.

To generate a random face with the same segmentation mask,
we 
\revAA{first note}{note} 
that
also a
segmentation mask has a corresponding inverted latent code
\revAA{$w$ in the latent space $W$.}{$L_M \in W$.}  
Now, 
\revCC{by simply}{by} 
exchanging a portion of this latent code that is highly correlated to the identity, we can create a rough effect of identity swapping. We have found that using this simple trick allows the segmentation mask to be used as a proxy between the original face and the newly generated face, while still ensuring that the two faces have 
\revAA{exactly the}{the} 
same segmentation mask.  
\revAA{The main difference is that their identities are far apart and the generated faces can be made completely random within the boundary of the segmentation mask.}{} 

Formally, given a face $S$ and its latent code \revBB{$E(S)$}{$L_S = \mathcal{P}(S)$} in latent space $W$, we randomly sample a vector $z \in \mathcal{Z}$ and project it onto $W$ to get a latent code \revBB{$r=p(z)$, of a random face $R=G(r)$.}{$L_R = p(z)$ of a random face $R=G(L_R)$, where $p: \mathbf{Z} \to W$ is a mapper to map $Z$ space and $W$ space.}  
The segmentation mask $M_S$ 
\revCC{}{(extracted from $S$)} 
is then projected onto latent space $W$ to get the latent code $L_{M_S}=\mathcal{P}(M_S)$. The trick for generating a random face $R'$ that has the random identity $R$ and shares the segmentation mask $M_S$ with $S$ is to swap the components that have the highest identity disentanglement scores in \revBB{$r$}{$L_R$} onto $S$; i.e., \revBB{$r' = IDMask(S, r)$}{$L_{R'} = WSwapper(L_S, L_{M_S}) = L_S \cdot M_W + L_{M_S} \cdot (1-M_W) $}, where \revBB{$r'$ is masked latent code and $R'=G(r')$}{$WSwapper(.,.)$ is a swapping function that uses the identity layers/channels mask $M_W$ in $W$.
After that we can get random face $R' = G(L_{R'})$ sharing a segmentation mask with $S$}. Finally, 
we can generate an output face $O$ by selectively swapping the masked area $Seg(M_S)$ of choice between the source face $S$ and 
the newly generated face $R'$ 
\revAA{(that shares segmentation mask $M_S$ with $S$)}{(that share segmentation mask $M_S$)} as follows:
\begin{align}\label{eq:segmask}
    O & = SegSwapper_{Seg(M_S)}(S,R') \nonumber\\
      & = S \cdot Seg(M_S) + R' \cdot (1 - Seg(M_S)).
\end{align}

{\bf Tunable anonymity:}
To provide tunable anonymity and explore the best design tradeoffs, we consider both the impact of using different mask areas, each capturing different sets of identity-related features, and how the faces that are swapped are generated.

\subsection{Example results}

We next demonstrate the tunability of the approach.
First, Tab.~\ref{tab:mask-numbers} shows the identity distances when swapping different example areas. Here, we show both the mean and standard deviation over 30,000 example images in CelebAMask-HQ dataset~\cite{lee2020maskgan}.
Second, Fig.~\ref{fig:mask_varians} shows visual example results when masking on different areas of the face (top row).  Overall, we have found that the approach provides excellent control of the level of anonymization to be achieved and that the results nicely address the shortcomings of only doing swapping in latent space, making it more attractive for videos and other contexts that place stricter requirements on preserving specific facial features and/or the background.

The main drawback of the approach is that
\revBB{the it}{it} 
can result in a mismatch of lighting between the randomly generated face and the original face. While we have found that  most (but not all) such cases can be nicely corrected by a match color process, such a process requires additional care.  We have also found a few cases \revAA{}{when}
the segmentation mask was not well parsed, highlighting that the quality of the preservation depends on the accuracy of current state-of-the-art face parsers.  In the next section we show how these results can be used to automate the face generation process.

\begin{table}[t]
\centering
\caption{Mean and standard deviation of identity distance when swapping a specific area in the face.}\label{tab:mask-numbers}
\vspace{-8pt}
\resizebox{0.48\textwidth}{!}{%
\begin{tabular}{ c |c|c|c|c  }
& Base & Eyes & Eyes + Nose & Eyes + Nose + Mouth\\
 \hline
 ID distance   & 1.25 $\pm$0.10   &0.28 $\pm$ 0.08 &   0.72 $\pm$ 0.14 & 0.79 $\pm$ 0.14 \\
\end{tabular}}
\vspace{-6pt}
\end{table}

\begin{figure}[t]
\centering
\includegraphics[width=\linewidth]{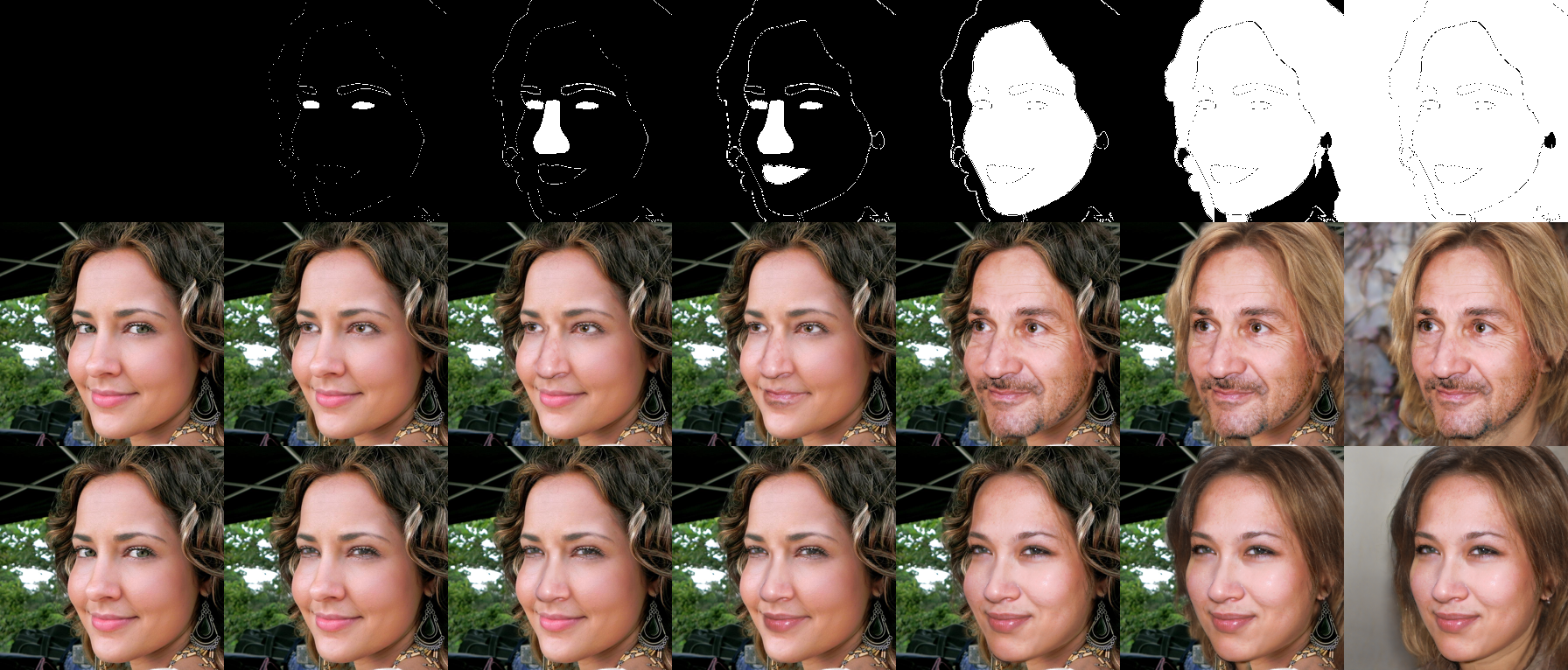}
\vspace{-18pt}
\caption{Example results when masking on different areas of the face (top row). 
\revAA{From left to right we see (1) no masking, (2) masking the eyes only, (3) masking eyes+nose, (4) masking eyes+nose+mouth, (5) masking full facial area, (6) masking facial area and hair, and finally (7) full masking.}{Masked area from left to right: (1) none, (2) eyes only, (3) eyes+nose, (4) eyes+nose+mouth, (5) facial area, (6) facial area + hair, and (7) full masking.}}
\label{fig:mask_varians}
\vspace{-12pt}
\end{figure}

\section{Latent swapper}\label{sec:mapper}

In previous sections, we proposed and evaluated two basic and easy-to-use methods that help anonymize the identity in latent space and pixel space.
Both methods avoid the need to train big GAN models. This property is attractive since training large GANs are known to be complex and expensive (limiting who can train them). The models also provide desirable results for complementing problems.
However, both approaches have some shortcomings that we address next.  
In particular, 
\revAA{we would like to}{we present a method to} 
automate the process of anonymizing faces in latent space, while obtaining similar results to the segmentation-based 
\revAA{results achieved after fixing any potential lighting issues using a match color process.}{results.}  

\subsection{High-level approach}

To achieve our aim, we build a model that anonymizes a face in latent space using a ground truth built upon the results from our segmentation mask approach.
The input of the model is latent codes 
\revAA{that belong to}{of} 
a real face, and the 
\revAA{goal of the model is}{model is trained} 
to find an $\alpha$-mask that 
\revBB{hides}{pushes} 
the 
\revAA{input identity further away from}{identity sufficiently far away from source identity that it is not recognized using}
the identification threshold of the 
\revAA{facial recognition system.}{FRS.} 

As illustrated in Fig.~\ref{fig:swapper}, we divide the model into three sub-modules: (1) a coarse attributes swapper, (2) an identity swapper, and (3) a fine attributes swapper. 
\revAA{As analyzed in Sec.~\ref{sec:layers}, the middle group of layers holds most of the identity-related information of the latent codes. To help the swapper during training, the loss function therefore give a higher weight to this module.}{Since the middle layers (see Sec.~\ref{sec:layers}) hold most of the identity-related information, the loss function used during training gives most weight to the second module.}

\subsection{The loss function}
We use the results of the segmentation mask swapper from Sec.~\ref{sec:mask} as the expected ground truth for the latent swapper model.  This provides an anonymized face that shares the exact background and non-identity information but has a different identity.
During the training process, we use two inputs to the model: the source latent code 
\revBB{$s \in W$}{$L_S \in W$} 
and 
\revAA{}{the}
latent code 
\revBB{$r$}{$L_R$} 
of a random face $R=\mathcal{P}(L_R)$.
Finally, we 
\revCC{optimize the model}{minimize the following loss} 
so that it produces an identity mask $\alpha$ in latent space:
\begin{equation}
\mathcal{L} = \textrm{argmin}_{\alpha \in (0,1)}^{m \times n} \lambda_{L_2} | \revBB{s}{L_{\hat{S}}} - t_{truth} | \revDD{+}{-} \lambda_{ID} \mathcal{L}_{ID} (G(\revBB{s}{L_{\hat{S}}}),S ),
\end{equation}
where $\revBB{s}{L_{\hat{S}}} = \alpha \cdot \revBB{s}{L_S} + (1-\alpha) \cdot \revBB{r}{L_R}$; $m \times n$ is the size of latent codes in $W$; $\lambda_{L2}$ and $\lambda_{ID}$ are the lambda factor of the $L2$ latent loss and the cosine similarity between the identity of $G(\revBB{s}{L_{\hat{S}}})$ and $S$; $t_{truth}$ is the ground truth for the output of the model \revBB{which is the latent codes in $W$ of $T$ in equation (\ref{eq:segmask});}{ which is latent code of a target face $T$} and $\revBB{\mathcal{L}_{ID}(G(s),S)}{\mathcal{L}_{ID}(.,.)}$ is the identity loss.

The second part of the 
objective function 
\revEE{}{(eqn. (5))}
ensures 
that the anonymized face  $G(\revBB{s}{L_{\hat{S}}})$ and $S$ are further away in terms of identity distance. This is in fact the objective of the whole anonymization process. 
We see that the identity distance between $S$ and output $G(\revBB{s}{L_{\hat{S}}})$ gradually decrease 
due to there being pairs of $S$ and $R$ in the training dataset that are too close in term of identity distance.
Furthermore, $Dist_{ID} (S, R) < Dist_{ID}(S, T)$, since during the process generating $T$, we use some components from $S$ both in latent space and pixel space. 
Finally, we calculate the identity loss as:
\begin{equation}
\mathcal{L}_{ID}(\revBB{s}{S},\revBB{s}{\hat{S}})=C(F(G(\revBB{s}{L_{\hat{S}}})), F(S)),
\end{equation}
where $C(.,.)$ is the cosine similarity function and $F(.)$ is a facial recognition model that produces identity embedding vectors.

\subsection{Training and example results}
    We used Arcface~\cite{deng2019arcface} as the facial recognition model during the training. Our model is trained on CelebAMask-HQ dataset~\cite{lee2020maskgan} with 30,000 facial images in total, divided into training and testing sets by factor 90:10. Before the training process, we prepare the ground truth using the annotated segmentation mask provided in CelebAMask-HQ. Random faces are generated using 30,000 random vectors in latent space \revBB{$Z$}{$\mathcal{Z}$} projected to latent space $W$, and then passed through the generator to get the random faces. The process of swapping identity in the random face set and blending them into the source face is described in Sec.~\ref{sec:mask}. The swapper network has a similar structure to the mapper in StyleGAN but has only 4 fully connected layers. The input size is also double the size of latent code in $\mathcal{Z}$ so that it can take both some source latent codes $\revBB{s}{L_S}$ and latent codes $\revBB{r}{L_R}$ of random faces as input. Another important difference is that the swapper will pass (or put low weights) to layers that have low identity disentanglement, including layers [0,4] and [12,17]. Passing through these layers significantly boosts the network's convergence time. Finally, we train our model with hyperparameters $\lambda_{L2}= 1$, $ \lambda_{ID} = 0.1$, and learning rate 
    \revAA{$lr=0.1$}{0.1.}

\begin{figure}[t]
\centering
\includegraphics[width=0.68 \linewidth]{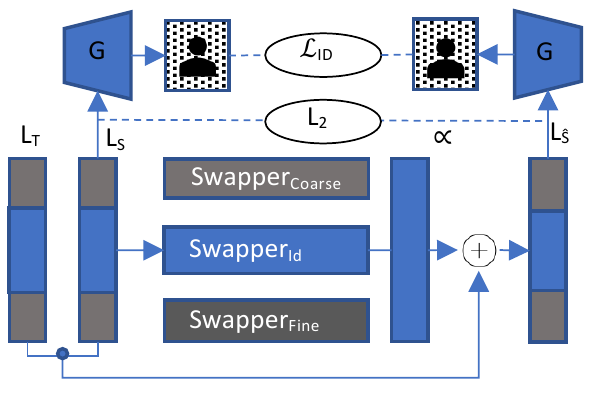}
\vspace{-10pt}
\caption{Network architecture of the latent swapper}
\label{fig:swapper}
\vspace{-18pt}
\end{figure}

Fig.~\ref{fig:swapper_result} shows example results. The results highlight that the mapper produces results (bottom row) 
\revAA{that nicely push the identity}{where the identity has been pushed} 
towards the semi-randomly generated face (second row) while preserving many of the features of the source 
\revAA{faces}{face} 
(top row).  We next evaluate the performance of our system 
\revAA{}{(Sec.~\ref{sec:evaluation})}
and compare the results with those produced by related 
\revAA{works.}{works (Sec.~\ref{sec:related}).} 

\begin{figure}[t]
\centering
\rotatebox{90}{\parbox{5.8cm}{\phantom{-------}{\small Swapper} \phantom{---------} {\small Rand.mask} \phantom{--------} {\small Source} }}
\includegraphics[width=0.86\linewidth]{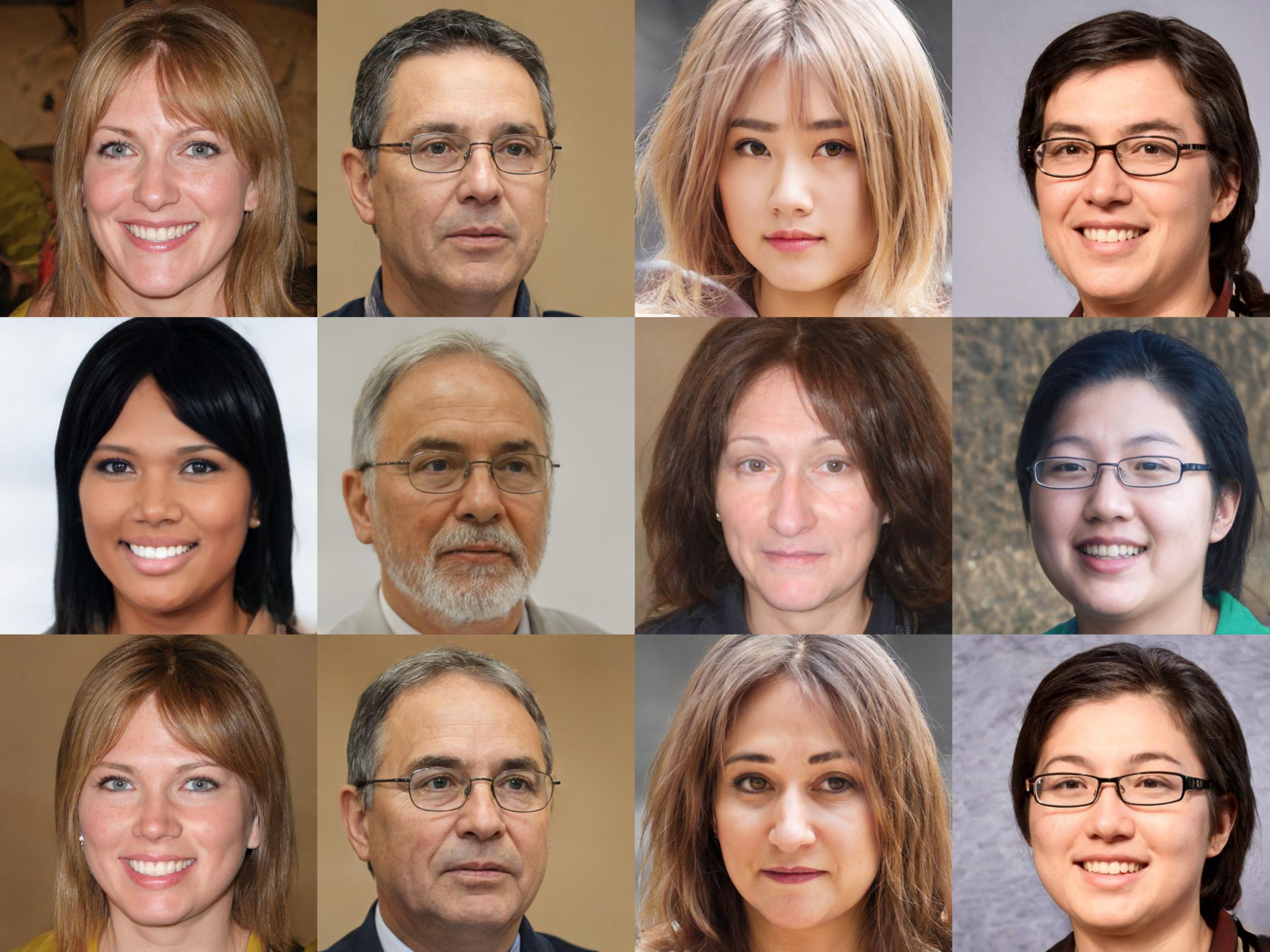}
\vspace{-10pt}
\caption{Example result using the latent swapper. Source face (top), random face with same face mask and desirable distance (middle), and output 
\revAA{face generated by swapper}{face}
(bottom). \revBB{}{Attributes to preserve: age, gender, face expressions, glasses.}}
\label{fig:swapper_result}
\vspace{-8pt}
\end{figure}

\begin{table*}[t]
\centering
\caption{Comparison of identity distance before and after applying different anonymization/de-identification methods.}\label{tab:id-dist}
\vspace{-10pt}
\begin{tabular}{ c|c|c|c|c|c|c  }
 FR model& Baseline & StyleID & DeepPrivacy~\cite{hukkelaas2019deepprivacy} & 
 k-same~\cite{newton2005preserving} & 
 AnonFACES~\cite{le2020anonfaces} & 
 Fawkes~\cite{shan2020fawkes}\\
 \hline
 FaceNet   & 1.03 $\pm$ 0.08 &   1.18 $\pm$ 0.08 & 1.2 $\pm$ 0.32 & 0.89 $\pm$ 0.12 & 0.98 $\pm$ 0.11 & 0.65 $\pm$ 0.08 \\
 ArcFace   & 1.19 $\pm$ 0.070 &   1.35 $\pm$ 0.11 & 1.21 $\pm$ 0.40 & 0.82 $\pm$ 0.13 & 0.97 $\pm$ 0.09 & 0.62 $\pm$ 0.06 \\
 CurricularFace  & 1.22  $\pm$ 0.13 &   1.29 $\pm$ 0.08 & 1.29 $\pm$ 0.42 & 0.98 $\pm$ 0.20 & 0.10 $\pm$ 0.11 & 0.72 $\pm$ 0.13 \\
\end{tabular}
\vspace{-6pt}
\end{table*}

\begin{figure*}[t]
\centering
\begin{subfigure}{.3\textwidth}
    \centering
    \includegraphics[width= 0.86\linewidth]{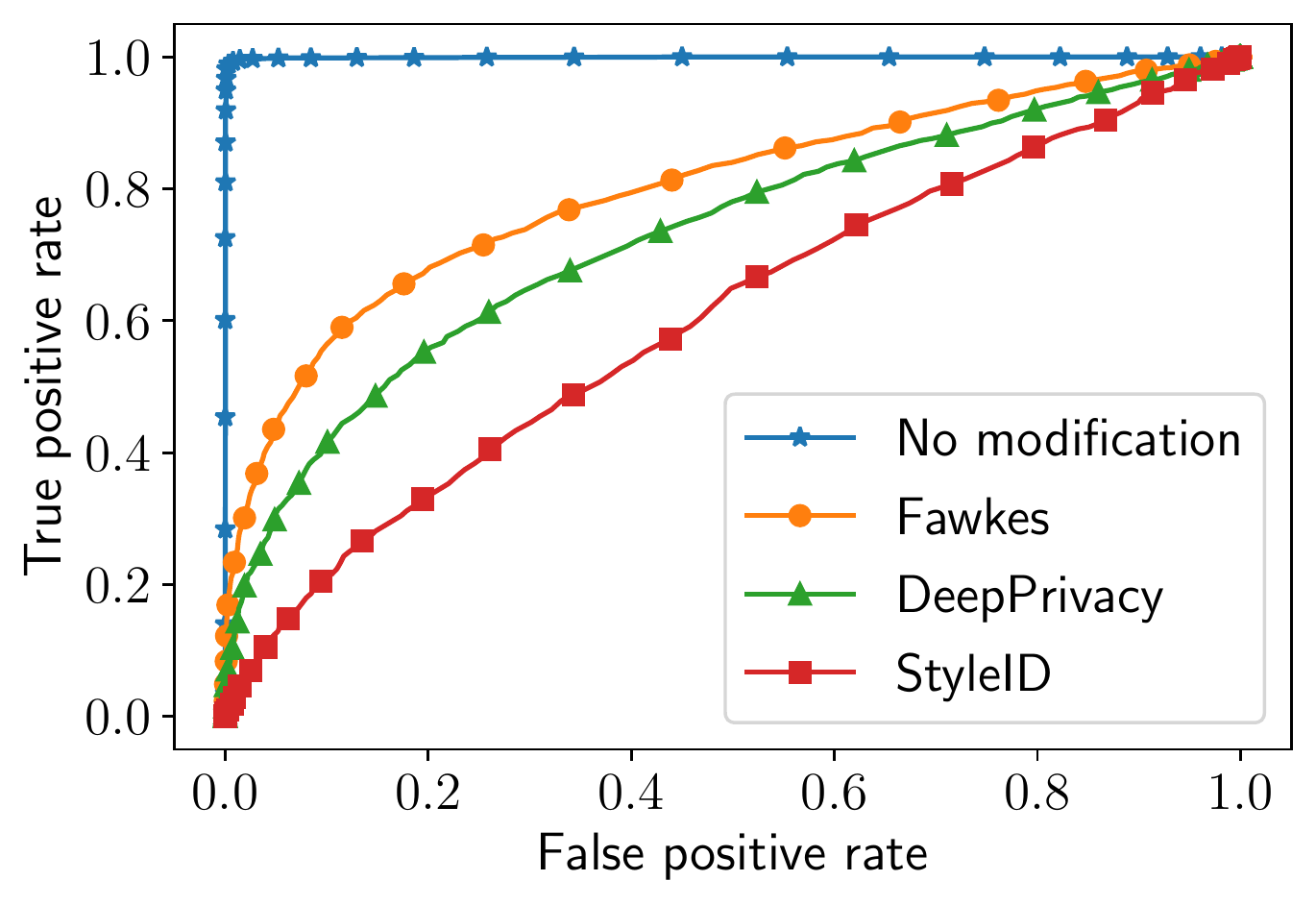}
    \vspace{-8pt}
    \caption{FaceNet.}
    \label{fig:facenet_lfw}
\end{subfigure}
\begin{subfigure}{.3\textwidth}
  \centering
    \includegraphics[width= 0.86\linewidth]{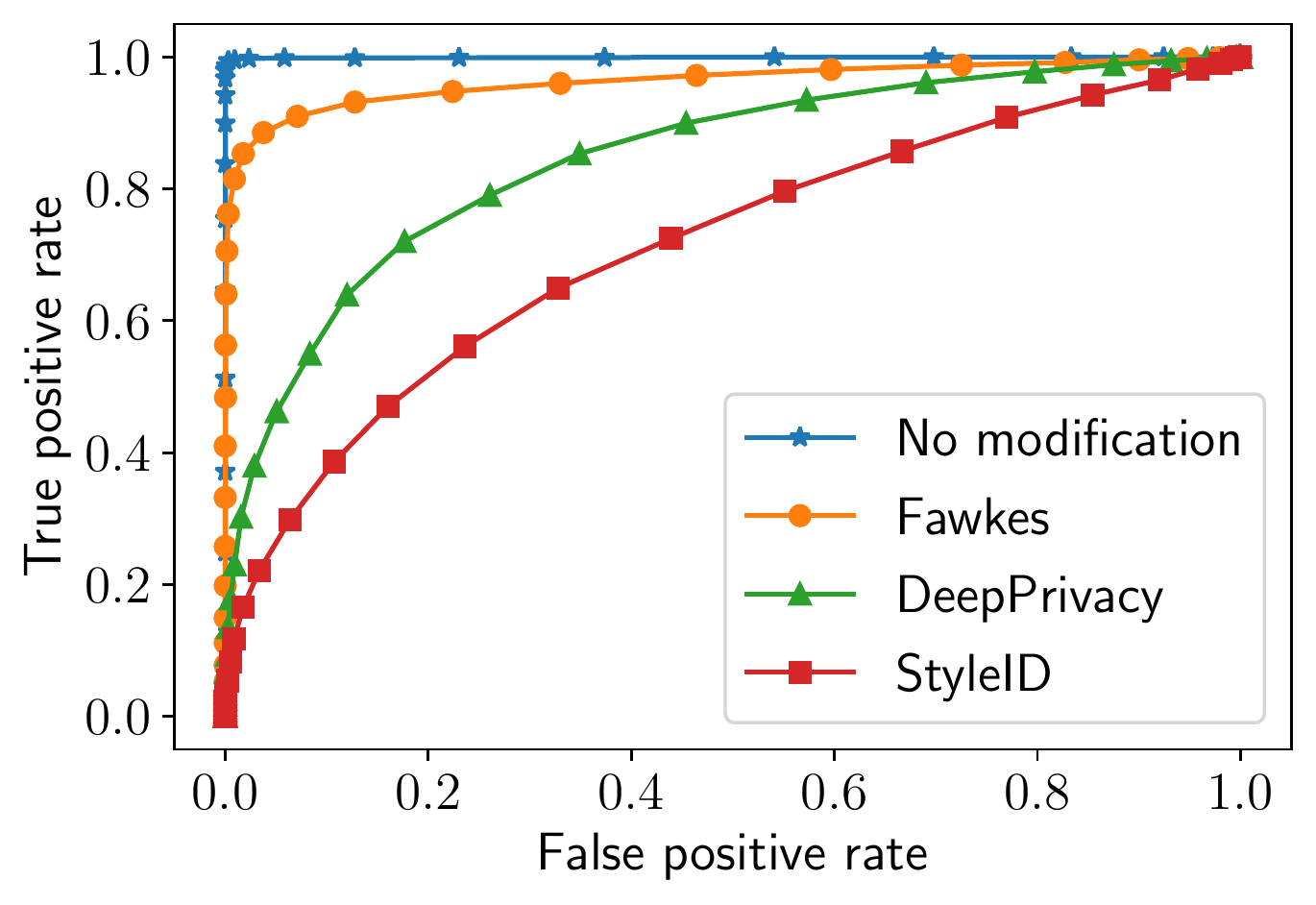}
    \vspace{-8pt}
    \caption{ArcFace.}
    \label{fig:arcface_lfw}
\end{subfigure}
\begin{subfigure}{.3\textwidth}
  \centering
    \includegraphics[width= 0.86\linewidth]{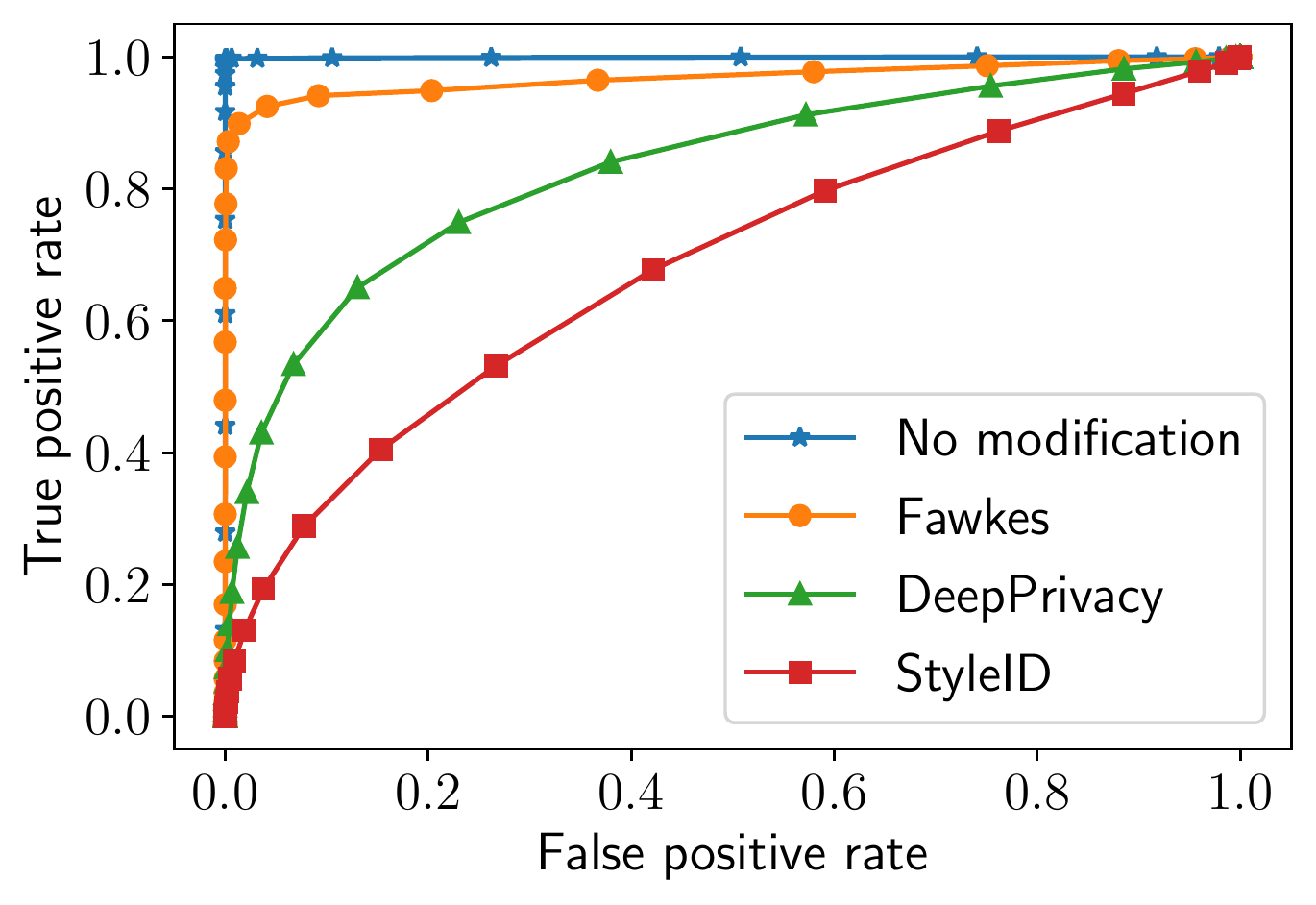}
    \vspace{-8pt}
    \caption{CurricularFace.}
    \label{fig:curri_lfw}
\end{subfigure}
\vspace{-12pt}
\caption{\revAA{}{ROC curves of different face embedding models on LFW benchmark.}}
\label{fig:roc_fig}
\vspace{-10pt}
\end{figure*}

\section{Evaluation}\label{sec:evaluation}

\subsection{Privacy vs. facial recognition system \revAA{}{(FRS)}}\label{sec:eval-frs}

We first evaluate 
\revEE{our frameworks}{StyleID's}
capability to protect the anonymized face against facial recognition and compare the results we achieve (here represented using the latent swapper) with the performance of other anonymization and de-identification methods. 
\revAA{}{To prevent the 
``data leakage” problem,
we use another dataset for evaluation.} 
For 
\revBB{evaluation}{privacy evaluation against FRS} 
we use the 
\revAA{LFW dataset~\cite{huang2008labeled}, which includes 12,433 images of 5,749 identities.}{LFW dataset~\cite{huang2008labeled}. The LFW is a standard dataset commonly used for evaluating the Facial Recognition models. The dataset includes 12,433 images of 5,749 identities and comes with a protocol to evaluate the facial recognition accuracy.} 
We report results for 
\revBB{two}{three} 
metrics. 
\revCut{
The first metric is the Euclidean distance between facial embeddings of the original source faces and the corresponding anonymized faces.  
To calculate the embeddings we use three state-of-the-art facial recognition models: Facenet~\cite{schroff2015facenet}, Arcface~\cite{deng2019arcface}, and CurricularFace~\cite{huang2020curricularface}. 
The 
second
metric we 
us
is the average of the true 
distance rank
in the gallery dataset.
Here, 
distances
are calculated by passing the anonymized faces through 
a facial recognition solution
built on Microsoft face 
API.
The custom built facial recognition has gallery and probe sets that are splits from the LFW dataset with a ratio of 90:10.
}

\begin{table}[t]
\centering
\caption{\revAA{}{Area under the ROC curve (AUC) after applying different anonymization/de-identification methods.}}\label{tab:auc}
\vspace{-10pt}
\resizebox{0.47\textwidth}{!}{%
\begin{tabular}{ c|c|c|c|c  }
 FR model& Baseline & StyleID & DeepPrivacy~\cite{hukkelaas2019deepprivacy} & 
 Fawkes~\cite{shan2020fawkes}\\
 \hline
 FaceNet   & 0.9994 &   0.6011 & 0.7291 & 0.7983 \\
 ArcFace   & 0.9994 &   0.7127 & 0.8465 & 0.9636 \\
 CurricularFace  & 0.9994 &   0.6805 & 0.8310 & 0.9684\\
\end{tabular}}
\vspace{-14pt}
\end{table}

\begin{table}[t]
\centering
\caption{\revAA{}{Accuracy on LFW benchmark when applying different anonymization/de-identification methods.}}\label{tab:acc}
\vspace{-10pt}
\resizebox{0.47\textwidth}{!}{%
\begin{tabular}{ c|c|c|c|c  }
 FR model& Baseline & StyleID & DeepPrivacy~\cite{hukkelaas2019deepprivacy} & 
 Fawkes~\cite{shan2020fawkes}\\
 \hline
 FaceNet   & 0.9935 &   0.5755 & 0.6775 & 0.7347 \\
 ArcFace   & 0.9955 &   0.6599 & 0.7718 & 0.9238 \\
 CurricularFace  & 0.9981 &   0.6325 & 0.7538 & 0.9411\\
\end{tabular}}
\vspace{-14pt}
\end{table}

\revAA{{\bf Distance-based comparison:}}{{\bf 1) Distance-based comparison:}}
\revAA{}{The first metric is the Euclidean distance between facial embeddings of the original source faces and the corresponding anonymized faces.  
To calculate the embeddings we use three state-of-the-art facial recognition models: Facenet~\cite{schroff2015facenet}, Arcface~\cite{deng2019arcface}, and CurricularFace~\cite{huang2020curricularface}.}
Tab.~\ref{tab:id-dist} presents results using 
\revAA{the first}{this} 
metric.  Here, we show the average identity distance before and after applying different anonymization/de-identification techniques (plus/minus the standard deviation). As a reference, we also include a baseline column that contains the mean and standard deviation of the pairwise Euclidean distance between random pairs of faces (belonging to different identities) in the evaluation dataset. 

To avoid the risk of re-identification, it is desirable that the distance is equal or larger than 
\revBB{for the}{the} 
baseline.  Both our framework and DeepPrivacy~\cite{hukkelaas2019deepprivacy} achieve distances well above this baseline for all three face recognition tools.  The very good performance using our framework when evaluated against ArcFace can be explained by our SwapperNet being optimized
\revBB{with using}{using the} 
ArcFace identity 
embedding during the training process. Another noticeable difference is that DeepPrivacy has much higher standard deviation than us.  One reason for this difference is that 
\revAA{the}{their} in-painting method 
has some levels of entanglement with the surrounding context, while there is no mechanism to ensure the newly generated face 
\revAA{are}{is} 
random. We observed this behavior while inputting two relatively similar faces and the output results are roughly the same. 

\revDD{Turning to face averaging methods such as k-same~\cite{newton2005preserving} and AnonFACES~\cite{le2020anonfaces}, we note that the average distance achieved by these methods are well below the baseline.}{The average distance of the two face averaging methods (k-same~\cite{newton2005preserving} and AnonFACES~\cite{le2020anonfaces}) are both well below the baseline.} 
This is because these methods average faces in a cluster into a single face. 
\revCut{Here it is important to note that this does not mean the outputs from these algorithms are at risk, just that they have higher risk of re-identification despite their k-anonymity property.} 
Finally, Fawkes~\cite{shan2020fawkes} has the lowest score using this 
\revAA{metric.}{metric, suggesting that its privacy filter sees the biggest chance of re-identification.} 
Part of the reason may be that 
\revAA{this method}{Fawkes} 
has not been optimized against state-of-the-art facial recognition models trained with triplet loss such as those that we are using. 
\revAA{The result shows that the chance of re-identification if using the privacy filter introduced in Fawkes is relatively high compared to 
\revBB{with the}{the} 
other methods.}{}

\revAA{}{{\bf{2) Receiver operating characteristic (ROC) comparison:}}}
\revAA{}{A popular approach
to evaluate how effective anonymization methods are against an FRS is to use ROC curves. 
\revCC{}{We use LFW's benchmark protocol~\cite{huang2008labeled}, which requires multiple face images per identity, and exclude
AnonFACE~\cite{li2019anonymousnet-extra} and k-same~\cite{newton2005preserving} (assumed one image per identity) from this evaluation.}
Figs.~\ref{fig:facenet_lfw}, \ref{fig:arcface_lfw} and~\ref{fig:curri_lfw} show head-to-head comparisons of our method when using FaceNet, ArcFace, and CurricularFace.  In all cases, we compare against both Fawkes~\cite{shan2020fawkes} and DeepPrivacy~\cite{hukkelaas2019deepprivacy}. 
As a baseline we also include the use of the original dataset (``no modification" shown in blue). 
StyleID (red) significantly outperforms Fawkes (orange) and DeepPrivacy (green), as demonstrated by its ROC curves consistently being 
closer to the diagonal.
While Fawkes (orange) provides some protection against FaceNet, it provides very limited protection against the two newer face embedding models.
DeepPrivacy (green) consistently outperforms Fawkes but has 
less attractive tradeoff curves than StyleID.
Supporting these ROC results, we provide additional quantitative metrics: Area Under the Curve (AUC) shown in Tab.~\ref{tab:auc} and the Accuracy shown in Tab.~\ref{tab:acc}. Here, lower values are better.}

\begin{table*}[t]
\caption{Comparison of identification rank before and after applying different anonymization/de-identification methods}\label{tab:id-rank}
\vspace{-10pt}
\centering
\begin{tabular}{ p{2.5cm}|c|c|c|c|c|c  }
 & Baseline & StyleID & DeepPrivacy~\cite{hukkelaas2019deepprivacy} & 
 k-same~\cite{newton2005preserving} & 
 AnonFACES~\cite{le2020anonfaces} & 
 Fawkes~\cite{shan2020fawkes}\\
 \hline
 Identification rank   &  1.02 $\pm$ 0.08 &  1,027.21 $\pm$ 25.68 & 925.43 $\pm$ 107.54 & 107.32 $\pm$ 57.47 & 153 $\pm$ 32.51 & 5.29 $\pm$ 2.68 \\
\end{tabular}
\vspace{-12pt}
\end{table*}

\revAA{{\bf Rank-based comparison:}}{{\bf 3) Rank-based comparison:}}
\revAA{}{Third, we used the average of the true rank in the gallery dataset. Here, the ranks are calculated by passing the anonymized faces through an FRS built on Microsoft face API~\cite{msfaceapi}.}
\revAA{The evaluation using Microsoft's Face API}{Specifically, the evaluation} 
was carried out as follow:  (1) 
\revBB{LWF}{LFW} 
is split into gallery and probe sets. (2) The images in the gallery, with their identify labels, are used for 
\revBB{training}{building} 
the facial recognition. (3) For each probe image passing through the facial recognition model, the rank of the true identity is recorded. 
Here, we say that a probe image $p_i$ 
\revAA{(with identity label $id_i$) achieves}{achieves} 
rank $k$ when there are $k-1$ other identities that 
\revBB{the facial recognition tool estimates to be more likely to be}{have lower identity distance than} 
the identity 
\revBB{shown in}{in} 
the probe image $p_i$.
(4) The same procedure as in step 3 is applied on anonymized/de-identified images in the probe set. 
Tab.~\ref{tab:id-rank} reports the average and standard deviation over the whole probe set. 

Similar to the 
\revBB{previous}{first} 
evaluation, we also include a baseline where images from the same person are passed through the 
\revAA{facial recognition system.}{FRS.} 
The result shows that the Microsoft face's API is highly accurate in recognizing faces in the LFW dataset. We observe a similar trend to the identity distance metric.  
\revAA{Our framework}{StyleID} 
\revBB{achieve}{achieves} 
the best results, 
\revAA{which slightly}{slightly}
outperforming 
\revAA{DeepPrivacy~\cite{hukkelaas2019deepprivacy} with this metric.}{DeepPrivacy~\cite{hukkelaas2019deepprivacy}.}  
Both frameworks achieve an average rank of around 1,000 using a dataset with 
\revBB{roughtly}{around} 5,000 identities in the 
\revAA{gallery dataset.}{gallery.}  
This is 
\revBB{roughly}{approximately} 
6-10 times 
\revAA{as good as}{better than}
k-Same and AnonFACES, and 20 times better than 
\revAA{with Fawkes.}{Fawkes.}  

While the 
\revAA{above anonymization}{rank-based} 
results may suggest that our performance is only slightly better than DeepPrivacy~\cite{hukkelaas2019deepprivacy}, we note that our framework typically is much better at preserving facial features, and provides more 
\revShort{visually appealing}{appealing} 
and natural looking 
\revShort{results.  A visual example compassion is presented in Sec.~\ref{sec:related}.}{results (cf. Fig.~\ref{fig:multi_cmp}).}

\subsection{Attributes preservation }\label{sec:preserving_attr}

\begin{figure}[t]
\centering
\includegraphics[width=0.76\linewidth]{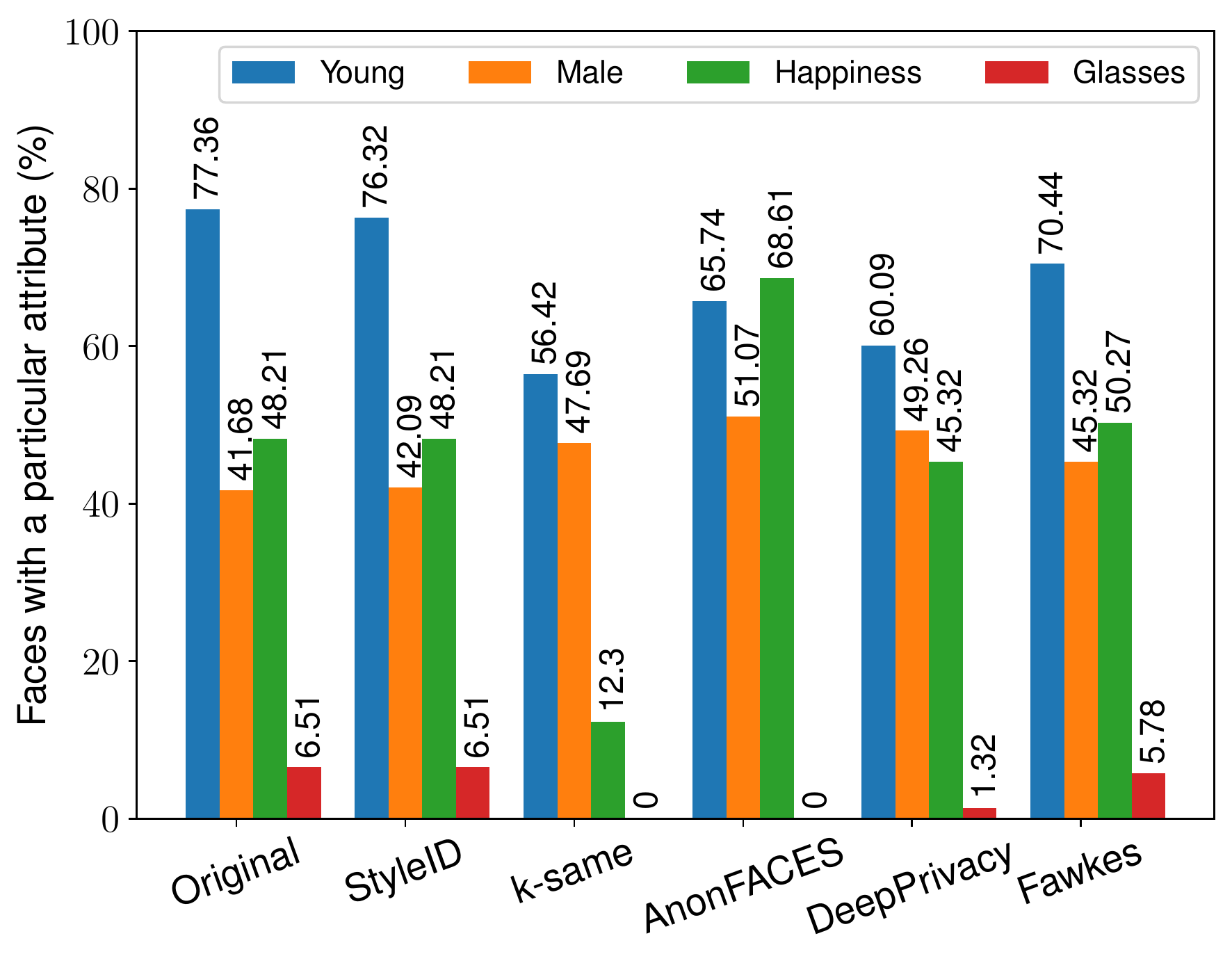}
\vspace{-16pt}
\caption{Attribute preservation. Percentage of faces 
with 
selected example attributes before (Original) and after applying different 
methods \revCC{}{(closer to ``Original" is better).}}
\label{fig:attr}
\vspace{-12pt}
\end{figure}

For high utility, 
\revAA{it is important that both the attributes of individual faces are preserved as well as 
the properties as summed up over all identities.}{it is important to preserve both the attributes of individual faces and the distribution of attributes in the dataset.}

{\bf Dataset properties:}
\revAA{Let us first consider}{We first consider} 
how well the methods preserve the distribution of example properties in the 
\revAA{dataset as a whole.}{dataset.}
This experiment is conducted on the CelebA dataset
\revBB{}{(we use the labelled attributes in the dataset as the attributes before anonymization)} 
and uses Microsoft's Face API as 
attribute 
\revBB{extractor.}{extractor after anonymization.} 
For this evaluation, we selected four example attributes of interest: ``Young" 
(defined as younger than 30 here), ``Male", ``Happiness", and ``Glasses". 
\revBB{The first three attributes are usually selected by the related works for this kind of evaluation while the last attribute is added to see if the algorithm can work well with a special attribute (this is a minority attribute in the dataset).}{The first three attributes are popular attributes in the dataset and the last 
is a minority attribute 
added to 
demonstrate that
the anonymizer 
works
well with 
special attributes too.} 

Fig.~\ref{fig:attr} shows the distributions of the four example attributes after applying the different methods 
\revBB{as well as for}{on}
the original image 
\revAA{dataset (bottom set).}{dataset.}  
We see that our framework provides a 
\revBB{very good}{good} 
match with the original 
\revAA{distribution.}{(bottom set).}  
Of the others, only Fawkes~\cite{shan2020fawkes} is able to preserve the minority attribute ``Glasses".  This is perhaps not surprising since Fawkes applies a very light privacy 
\revAA{filter (e.g., as seen by the worse protection against advanced facial recognition tools seen in the previous section).}{filter.}
The other techniques 
\revAA{do not do a good job preserving the properties,}{perform poorly,} 
often averaging out many attributes in individual faces, resulting in minority (attribute such as glasses often disappearing). 

\revAA{}{{\bf Controlling attributes of individual faces:}
While our default solutions are good at preserving many facial attributes, further control can easily be added.}
\revAA{Another advantage with manipulating the identity in 
the latent space is that we can control non-identity related attributes simultaneously.}{To show this, 
note that 
there exist many tools for controlling non-identity related attributes,
including StyleSpace~\cite{wu2021stylespace}, GANSpace~\cite{harkonen2020ganspace}, InterfaceGAN~\cite{shen2020interfacegan}, StyleCLIP~\cite{patashnik2021styleclip}.}
\revBB{}{In Fig.~\ref{fig:correct_gender}, we demonstrate a case where some randomly generated target faces (second row) have different gender compared to the original faces (first row). The gender is corrected in the third row by using global direction of GANSpace (only for faces with incorrect gender). In this case, our goal is to 
control the gender attribute of the target.
Referring back Sec.~\ref{sec:picking_layers}, the analysis suggested that swapping layers (5,6,7) will result in more attributes captured from the target. We therefore swap these layers of the latent codes in the first row, to the ones in the third row. The swapped results are shown in the fourth row in which the gender attribute is preserved.} 

\revAA{This is}{The properties outlined 
\revEE{in this section}{here} 
are} important for anonymization since 
\revAA{it allows}{they allow} 
us 
to add 
further prevention 
\revEE{against}{of}
re-identification attacks in which a set of meta-information about the face may be used to re-identify a person. 
\revAA{The}{For example, the} 
capability of the anonymization method to manipulate a given attribute is 
\revAA{crucial}{valuable} 
for concealing minority identities. 
While we 
\revAA{here have}{have} 
demonstrated that we can successfully preserve these properties, \revAA{we note that it is}{it is} 
trivial to change some subset of minority properties if 
\revDD{so}{this} 
would be desired (e.g., due to 
\revAA{the}{an} 
attacker having knowledge about all individuals' attribute values) by manipulation in latent space.  This allows us to easily further enhance the level of anonymization 
\revAA{by applying 
l-diversity
or t-closeness on attributes,}{(e.g., by applying l-diversity or t-closeness on attributes),} 
and by giving a constraint on the attributes before the anonymization process.

\begin{figure}[t]
\centering
\rotatebox{90}{\parbox{5cm}{\phantom{---}{\small Results} \phantom{----}{\small Corrected} \phantom{---} {\small Target} \phantom{----} {\small Source} }}
\includegraphics[width=0.96\linewidth]{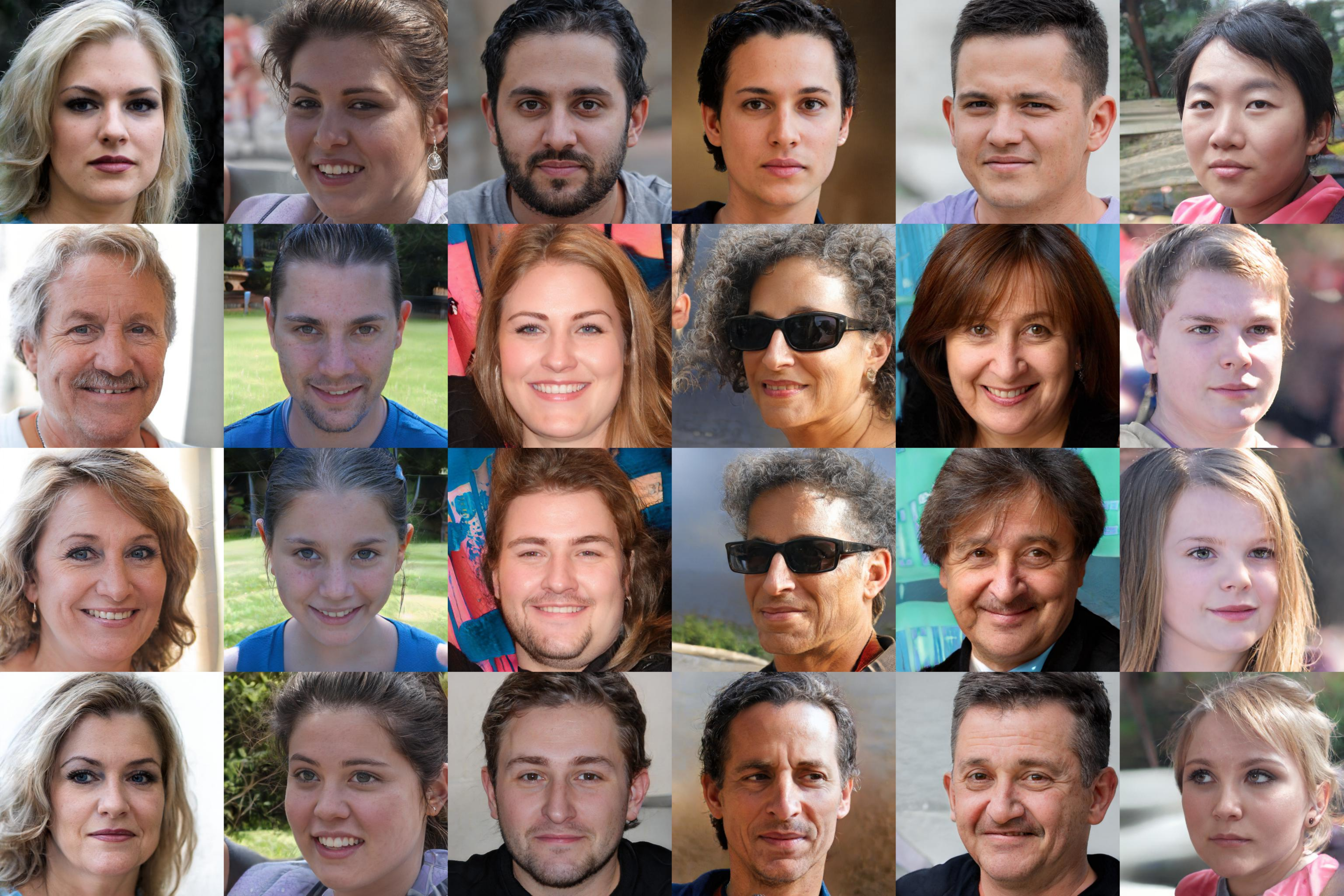}
\vspace{-18pt}
\caption{Controlling gender attribute in latent space. Top row: original faces, second row: random target faces, third rows: face with corrected gender attribute, bottom row: results by swapping layers (5,6,7)}
\label{fig:correct_gender}
\vspace{-12pt}
\end{figure}

\begin{figure}[t]
\centering
\includegraphics[trim= 0mm 18mm 0mm 16mm, clip, width=0.74\linewidth]{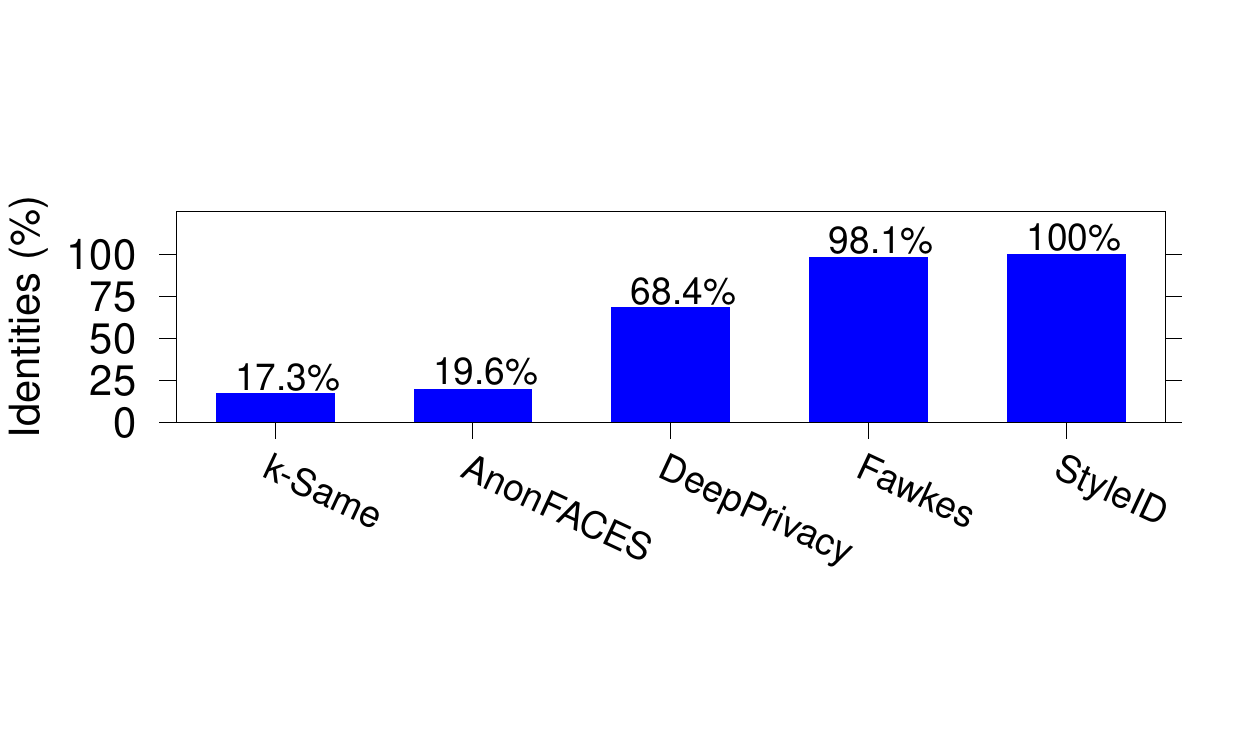}
\vspace{-12pt}
\caption{\revAA{Percentage of}{Relative} number of identities in the face image dataset after the anonymization/de-identification process.}
\label{fig:id_divert}
\vspace{-12pt}
\end{figure}

\subsection{Identity diversity}

\revShort{In addition to the attribute distributions (previous subsection), it is also important to preserve the number of identities in the dataset. This factor is important to ensure high utility of the dataset when training robust models for face detection, recognition, and generative models, for example.}{To ensure high utility, it is also important to preserve the number of identities in the dataset.}  
For this evaluation, 
\revAA{we again use the CebebA dataset~\cite{liu2015faceattributes} and extract a sample image from this dataset for each of the 10,777 identities in the dataset.}{we extract a sample image for each of the 10,777 identities in CebebA~\cite{liu2015faceattributes}.} 
\revShort{The evaluation process is executed as follow: 
(1) Anonymize/de-identify the images in the dataset using the different methods.
(2) Extract face embeddings from the result datasets using the CurricularFace model and cluster the embedding by the k-Mean clustering algorithm~\cite{lloyd1982least}. (3) The number of classes/ clusters is counted as the number
of identities in the anonymized/de-identified dataset. (4) Finally, we calculate the percentage of this identity number with the number of identities in the original dataset.}{After the anonymization/de-identification process of each method, we extract face embeddings using the CurricularFace 
\revAA{model and}{model,} 
cluster the embedding 
\revAA{by the k-means clustering algorithm~\cite{lloyd1982least},}{using k-means~\cite{lloyd1982least},} 
and use the number of clusters as a proxy for the number of identities in the final dataset. Fig.~\ref{fig:id_divert} shows the percentage of identities compared to in the original dataset.} 

\revAA{As shown in Fig.~\ref{fig:id_divert}, our capability to blend a random identity replacing the original identity while preserving most of the original image attributes allow us to preserve 100\% of the 
possible number of identities.}{As desired, StyleID preserves the 
number of identities.} 
\revAA{For k-same and AnonFACES, we choose $k=5$ as suggested in those papers, in both cases reducing the number of identities below 20\%.}{In contrast, the number of identities decreases to below 20\% when using k-same and AnonFACES with $k=5$ (as suggested in those papers).} 
This clearly is a weakness of k-\revBB{anoymity}{anonymity}-based methods. 
Another interesting observation is that the generated identities \revBB{in}{by} DeepPrivacy have low diversity, in the sense that different original faces can share visually similar anonymized faces. As a result, it sees a 32\% drop in the number of unique 
\revAA{identities compared to in the original dataset.}{identities.}
\revAA{Part of this reduction may be explained by visual inspection of their results (cf. Fig.~\ref{fig:fb} and results in~\cite{gafni2019live}).  A closer look suggests that the nose area becomes relatively the same for almost all of their results. Fawkes, on the other hand, does a good job of preserving the number of}{Of the other techniques only Fawkes does a good job preserving the number of} 
\revShort{identities 
but may be helped by the method not pushing the identities far enough from the source identities to provide strong protection against advanced face detection algorithms.}{identities.}

\begin{figure*}[t]
\vspace{-14pt}
    \begin{minipage}[t]{0.69\textwidth}
        \centering
        \rotatebox{90}{\parbox{10.6cm}{\phantom{-----} {\small StyleID} \phantom{------} {\small AnonFACES~\cite{le2020anonfaces}} \phantom{---} {\small DeepPrivacy~\cite{hukkelaas2019deepprivacy}} \phantom{---} {\small Fawkes~\cite{shan2020fawkes}} \phantom{---} {\small Source}}}
        \includegraphics[width=0.97\linewidth]{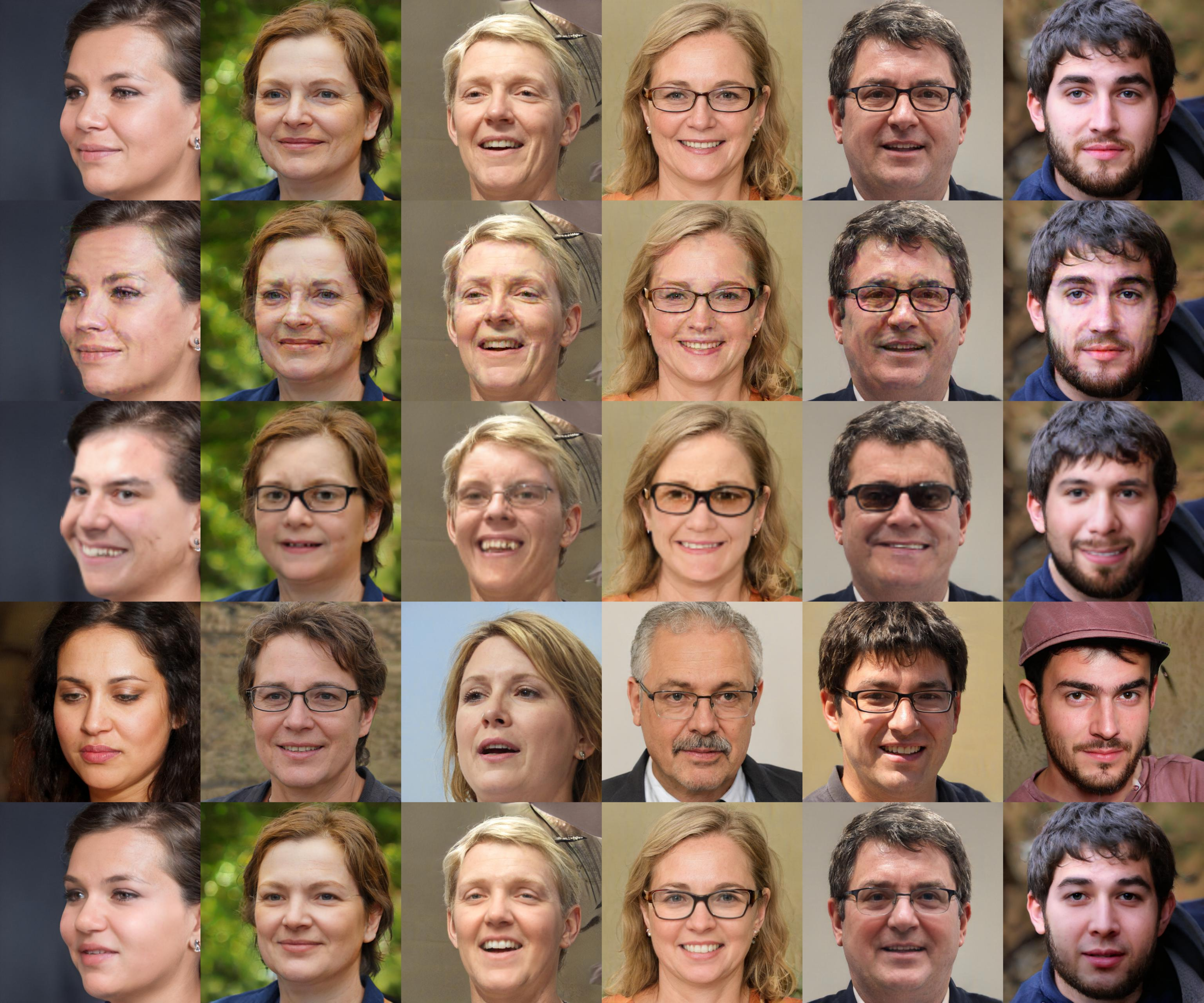}
        \vspace{-10pt}
        \caption{Comparison of our results to Fawkes~\cite{shan2020fawkes}, DeepPrivacy~\cite{hukkelaas2019deepprivacy}, AnonFACES~\cite{le2020anonfaces}}
        \label{fig:multi_cmp}
        \vspace{-2pt}
    \end{minipage}
    \hfill
    \begin{minipage}[t]{0.3\textwidth}
        \vspace{-282pt}
            \centering
            \rotatebox{90}{\parbox{4cm}{\phantom{-----}{\small StyleID} \phantom{-----} {\small Ref.~\cite{maximov2020ciagan}} \phantom{-----} {\small Source} }}
            \includegraphics[width=0.9\linewidth]{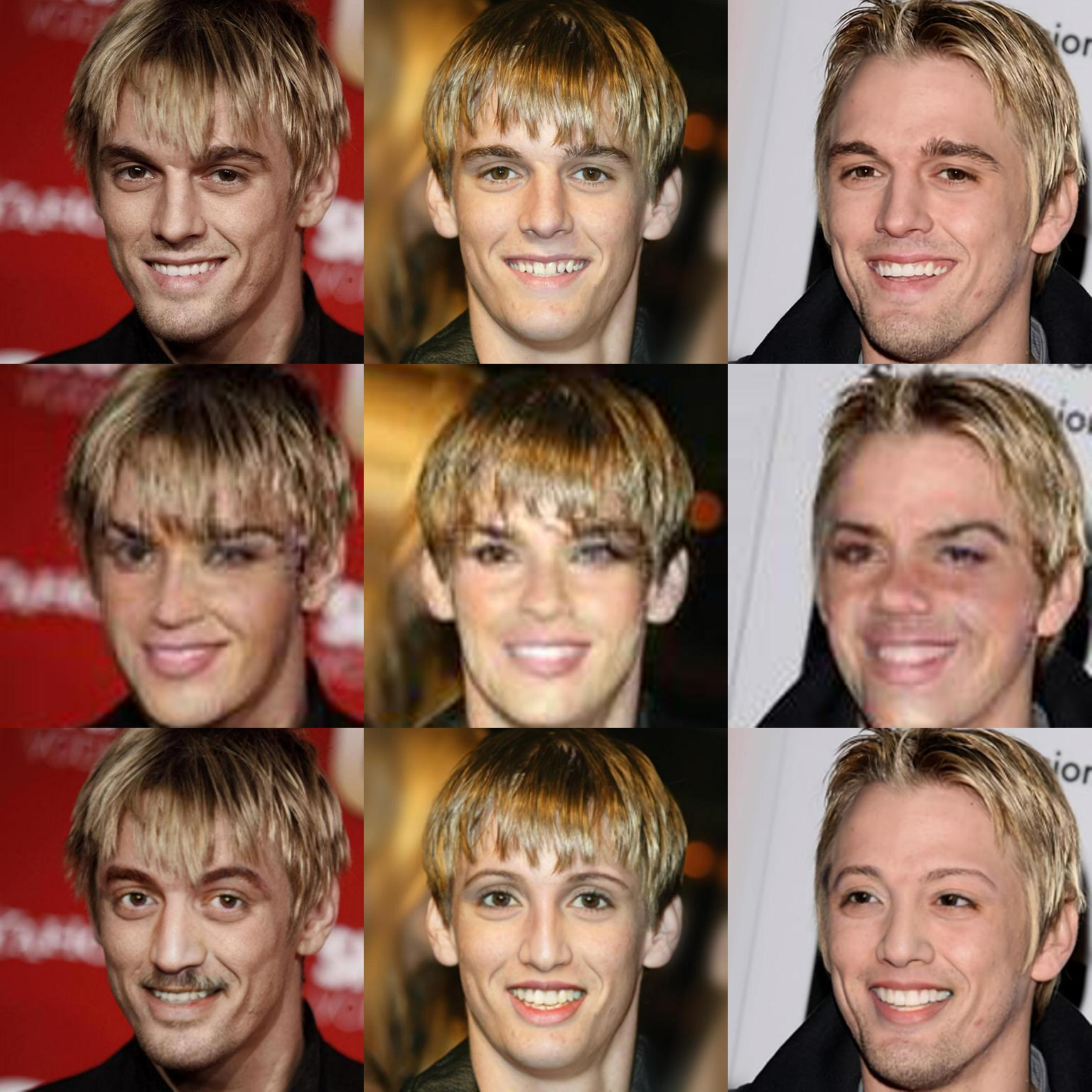}
            \vspace{-6pt}
            \caption{Comparison 
            CIAGAN~\cite{maximov2020ciagan}.}
            \label{fig:ciagan}
            \centering
            \rotatebox{90}{\parbox{4cm}{\phantom{-----}{\small StyleID} \phantom{------} {\small Ref.~\cite{gafni2019live}} \phantom{------} {\small Source} }}
            \includegraphics[width=0.9\linewidth]{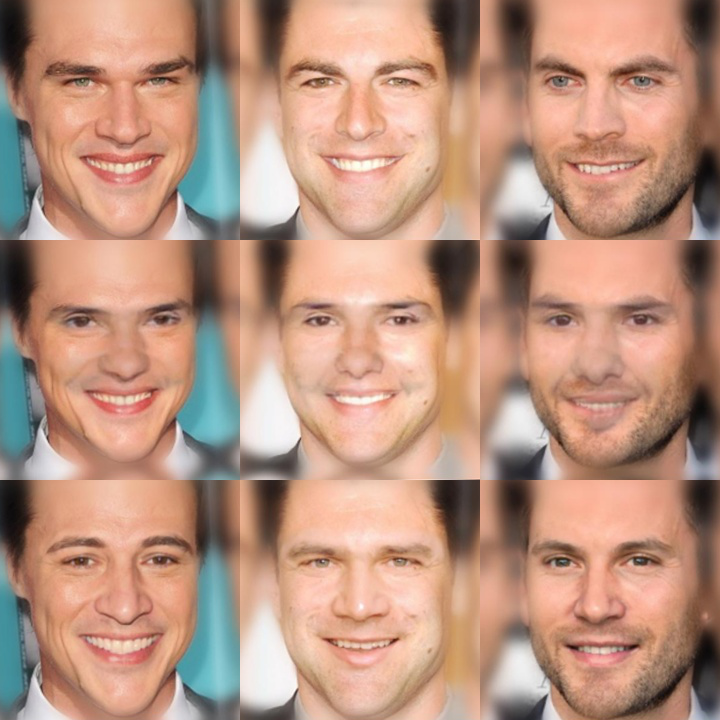}
            \vspace{-6pt}
            \caption{Comparison 
            Gafni et al.~\cite{gafni2019live}.}
            \label{fig:fb}
    \end{minipage}
    \vspace{-8pt}
\end{figure*}

\section{Comparisons with related works}\label{sec:related}

\subsection{Related work}

Early 
\revAA{works on anonymization of facial images}{anonymization works} 
were primarily based on the idea of k-Anonymity~\cite{sweeney2002k}.  
\revDD{Here, $k$ similar faces are clustered and replaced by the average face of the 
cluster. By doing so, the chance of re-linking an original face is at most $1/k$.
Works based on this idea is now referred to as k-same family algorithms and often
use different embedding spaces to find $k$ similar faces~\cite{meng2014retaining, sun2015distinguishable, du2014garp}.}{These works often use different embedding spaces to find $k$ similar faces~\cite{meng2014retaining, sun2015distinguishable, du2014garp}, and replace them by the average face of the cluster. This helps reduce the chance to re-link an original face to at most $1/k$.}  
For example, k-same-pixel~\cite{newton2005preserving} and k-same-M are based on active appearance model~\cite{gross2006model}, while AnonFACES~\cite{le2020anonfaces} recently showed how faces can be averaged in the latent space of a GAN. 
\revDD{Beside the 
\revAA{naturalness of the results,}{naturalness,} 
this method 
\revAA{posed}{poses} 
some limitations that makes it less suitable for practical use 
\revShort{cases.  For example, it does not translate well to video and (per definition) significantly reduces the number of unique identities in a dataset, limiting the scale of the datasets they can generate.}{cases (e.g., video and generation of large datasets).}}{The downsides of this approach are that it reduces the number of identities by a factor of $k$ and it does not consider attributes. We instead focus on privacy in the FRS context and derive privacy/utility metrics from the machine learning field.}

Recent works 
\revAA{on this topic have}{have} 
instead focused on identity de-identification with the sole purpose of concealing identity in facial images/videos. 
\revAA{For example, the works in}{For example,}
\cite{wu2019privacy,li2019anonymousnet,pan2019k,maximov2020ciagan,hukkelaas2019deepprivacy} all propose custom GAN models for image de-identification. However, they also all share the same weakness in that they lack naturalness. This is not a surprise since the facial domain is one of the most challenging domains for GAN models and the training process 
\revShort{requires a high cost both in term of computational resources and also the level of expertise to get a quality result.}{is complex and time consuming.} 

Gafni et al.~\cite{gafni2019live} apply similar models to faceswap~\cite{dfgithub} (originally known as deepfakes) to \revBB{de-identification in}{de-identify} video. This is one of a few solutions for videos. Unfortunately, there is no source code available and we could not re-produce the claimed results (but we can compare our \revBB{}{visual} outputs with theirs; see next section). 
Fawkes~\cite{shan2020fawkes} also aims to conceal the identity in facial images but takes a \revBB{somewhat}{}different approach (adds a privacy filter to the facial images) and the objective is \revBB{}{also} slightly different (i.e., adding invisible noises to hide identity in feature space of convolutional neural network). This way, they try to fool  
\revAA{some of the state-of-the art facial recognition systems}{some FRS:s} 
without significant changes in the pixel space.
The author provided the software and source codes so that anyone can test the results. However, on testing with real world image samples we found that this method is not effective against state-of-the art facial recognition models 
\revAA{which trained}{trained} 
with triplet loss such as FaceNet~\cite{schroff2015facenet}, ArcFace~\cite{deng2019arcface}, CurricularFace~\cite{huang2020curricularface}.

\subsection{\revBB{Comparing with GAN/CNN-based methods}{Visual comparison to the related work}}

\revAA{}{%
Fig.~\ref{fig:multi_cmp} compares StyleID with the main related works
that we could reproduce.
Here, the first row shows example of random source images.
For Fawkes (second row) we use a configuring of ``--mode mid", meaning that the protection is at middle level of three options (``low", ``mid", and ``high").  With this method, a form of random noise is added to the source images.
However, in contrast to the authors' claim of ``imperceptible" noise, the noise appears visible in our test results.
With DeepPrivacy~\cite{hukkelaas2019deepprivacy} (third row) we typically observed that the facial area had low resolution and looked un-natural.  Furthermore, we noted in Sec.~\ref{sec:preserving_attr}, attributes such as facial expression, glasses, and eyes direction were often not preserved.
AnonFACES~\cite{le2020anonfaces} (fourth row) achieved similar image quality as StyleID (it is also based on the StyleGAN generator) but did not have control of the facial attributes, often resulting in uncontrolled changes in attributes such as pose, expression, gender, and age.
The best performance was obtained with StyleID (fifth row), which achieved both high naturalness and nicely preserved the most important attributes (e.g., facial expressions, age, gender, glasses, etc.).} 

\revBB{In addition to DeepPrivacy~\cite{hukkelaas2019deepprivacy} (who provided the most comparable anonymization results in Sec.~\ref{sec:evaluation})}{In addition,} 
we selected to compare with CIAGAN~\cite{maximov2020ciagan} 
\revAA{(which provides a good baseline among works that have shared their code)}{(good baseline among works sharing their code)} 
and the work by Gafni et al.~\cite{gafni2019live} 
\revAA{(discussed above and which provides among the most promising results we had seen).}{(discussed above, provides among the most promising results we had seen, but do not share
their code).}  
Figs.~\ref{fig:ciagan} and~\ref{fig:fb} present comparisons with example results from these works.
In all cases our approach better preserve the facial features and naturalness simultaneously as we provide as good or better protection against 
\revAA{facial detection systems}{FRS:s} 
(e.g., see evaluation results in Sec.~\ref{sec:evaluation} for comparison with~\cite{hukkelaas2019deepprivacy}).
Similar to our original ideas, Gafni et al.~\cite{gafni2019live} use an identity loss to push the identity further away from original images. What we have found (and which is visible in their result) is that this \revCC{easily}{}can
\revBB{reduce}{converge} 
\revAA{the number of unique identities}{identities} 
as ``furthest away" may \revCC{be towards}{push results toward} the same point in 
\revAA{space for some feature.}{latent space.}  
\revBB{In their case, we suspect that this is the}{This is the}
reason almost all noses become 
\revCC{very similar.}{similar (see Fig.~\ref{fig:fb}).} 
To solve this problem (and ensure high visual diversity) we integrated the generation of random faces with a similar segmentation mask and then built that into the ground truth of the training of our latent swapper.
\revCC{We have also found that their results, depending on the input, in some cases fails to protect the identity against facial recognition. While there is a lambda factor on identity loss that can be used to adjusted, we were not able to reproduce their
\revBB{work (as the code was not shared)}{work}
to see to what extent that could solve the issue.}{} 
\revCC{In contrast, our result has}{This allows our results to have} 
a distinct look of a different 
\revAA{identity that not even the human eye should re-identified with
the original.}{identity.} 
Furthermore, the amount of 
\revAA{change on the result}{change} 
can 
\revAA{be easily}{easily be}
adjusted in our 
\revCC{case with the lambda factor,}{case,} 
as we do not require a lengthy re-training process (which~\cite{gafni2019live} would require). 

\section{Wider context}\label{sec:broader}
\subsection{\revAA{}{More advanced attacker model}}

\revAA{}{
In our evaluation in  Sec.~\ref{sec:evaluation}, we assumed an adversary that uses one of three different state-of-the-art FRS:s to try to re-identify the original face.  Here, StyleID was shown to significantly outperform prior works.  
A more advanced attacker may also have access to our anonymizer and may try to perform a type of parrot attack. Here, we assume a threat model where the attacker does not have access to the gallery set of the FRS but can create an arbitrary probe set, can make a large number of queries via the FRS's API, and can generate anonymized faces using the anonymizer (that can be feed to the FRS).
For this scenario, our use of random vectors to generate random synthetic faces ensures that the attacker cannot easily reverse engineer the original faces through guessing. 
This would be trivial if the model was deterministic.
Using a probabilistic function, the chance of the attacker guessing the correct anonymized face is as good as predicting the next random $z \in \mathcal{Z}$.  While the subset of $z \in \mathcal{Z}$ that corresponds to real faces is substantially smaller than the size of $\mathcal{Z}$, the probability of guessing the next face being generated is still small, making this a non-trivial task. In this discussion we have ignored that the attacker may use the background to identify an image from the probe set. To protect also against such attacker modifications also to the background would be needed. This is outside the scope of this paper.  In fact, to provide good utility we prefer to keep the background and hair in most cases.}

\subsection{\revDD{}{Generated faces match real people}}
\revDD{}{Using StyleGAN's face generator, we acknowledge that there is a risk that a generated face matches the face of a real person. This is an issue discussed in the GAN research community that follow-up works should be aware of. However, we note that the possibility of the generated face matching a real person is similar to finding two people with the same facial identity. In practice, with 8 billion people alive today, having a matching identity with another person is today not seen as privacy concern.}

\subsection{\revAA{}{Ethical statement}}

\revBB{}{Synthetic data generated by GAN models and DeepFakes can be misused for harmful purposes such as impersonation and 
misinformation. 
Our work intends
to protect facial identities and to benefit the machine learning community by enabling privacy-preserving collection and sharing of training data. 
\revDD{We will}{We} 
publish our source codes for transparency purposes and encourage the 
open-source 
community to build upon our work for the greater good.
However, we also acknowledge that the work, as well as the related works in the field of synthetic data generation, could be misused. We therefore raise a warning and highly recommend having the consent of the people in the collected images/video
and carefully consider the legal aspect before using the proposed techniques.}

\subsection{\revAA{}{Practical applications}}

\revAA{}{
StyleID provides an effective way to anomymize/ de-identify image datasets.  This helps protect individual's 
identity.
This has several applications, including to anonymize faces before uploading images to social media platforms or to create anonymized facial image data that can be used for research and/or to improve the accuracy and robustness of machine learning models (e.g., face analysis/recognition models, facial image synthetic models, 3D facial reconstruction models) applicable for a broad range of topics.
\revDD{}{StyleID may be particularly valuable when wanting to simultaneously preserve contextual information and respect the privacy of bystanders~\cite{hasan2020automatically} incidentally captured in photos.}
The insight from our work on identity disentanglement can also help develop DeepFake detection 
\revDD{algorithms.}{algorithms (e.g., leveraging our approach to swap the identity of a face to many random identities can add fidelity to their training datasets).} 
Finally, StyleID can easily be combined with other frameworks on semantic image editing in ways that could benefit fields such as photography, cinematic, and gaming.}

\section{Conclusion}\label{sec:conclusions}

\revAA{This paper has presented}{We have presented} 
the design and evaluation of 
\revAA{}{StyleID,}
a feature-preserving anonymization framework that protects the
\revAA{individuals’}{individual's}
identity, 
while preserving as many characteristics of the original faces in the image dataset as possible.  
\revAA{The framework}{StyleID} 
is based on the insights derived through the definition of a new identity disentanglement 
\revAA{}{metric}
and the incremental development (and evaluation) of three complementing disentanglement methods that each build upon the insights and results of the prior methods. 
\revAA{The framework}{StyleID} 
provides tunable privacy, 
\revBB{has low computational complexity,}{efficacy,} 
and is shown to outperform current state-of-the-art solutions.  The high utility of datasets that can be generated using 
\revAA{our framework are}{StyleID is} 
believed to be an enabler for more and better training data for machine learning models that need to operate in contexts 
\revAA{were}{where} 
there are people. 
\revDD{Our code will be shared with the final version of the paper.}{Our code can be found here:\\ 
{\url{https://github.com/minha12/StyleID}.}}

\begin{acks}
This work was supported by the Wallenberg AI, Autonomous Systems and Software Program (WASP) funded by the Knut and Alice Wallenberg Foundation.
\end{acks}

\bibliographystyle{ACM-Reference-Format}
\bibliography{reference}

\begin{figure*}[t]
\centering
\includegraphics[width=0.7\linewidth]{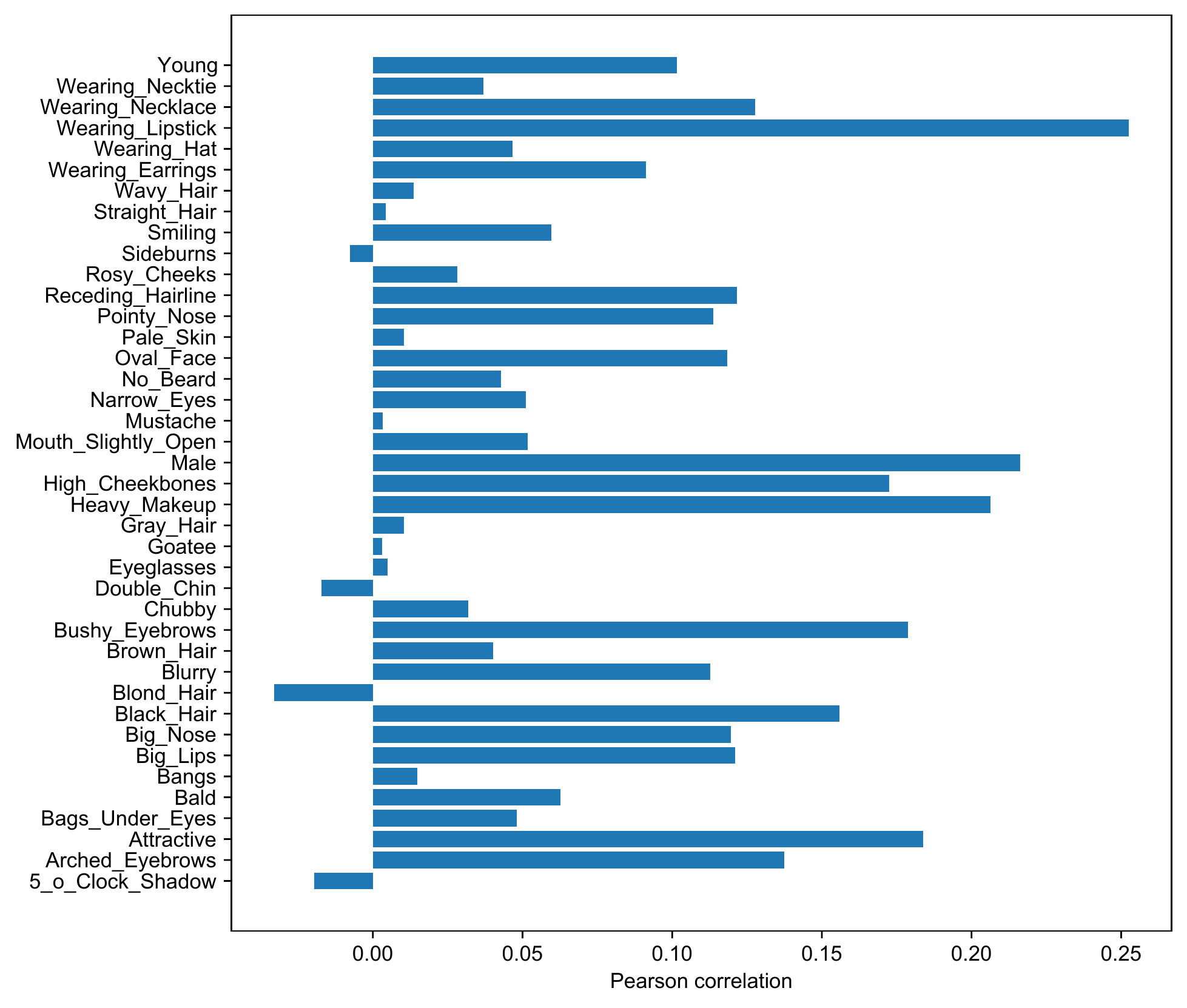}
\vspace{-10pt}
\caption{Correlation of change in identity distance and L1 distance of attributes when swapping layer [5,6,7]}
\label{fig:id_corr}
\vspace{-2pt}
\end{figure*}

\appendix

\section{Correlation between identity and individual attributes}\label{appd:A}

\revBB{}{The latent space of StyleGAN is among one of the most disentangled latent spaces in the sense that it allows manipulation of a particular attribute or a set of attributes while keeping the remaining attributes un-changed. As shown in the related works~\cite{patashnik2021styleclip,harkonen2020ganspace,shen2020interfacegan,wu2021stylespace} this allows meaningful manipulation of the attributes. However, when it comes to identity, we observed that there is a degree of correlation between identity and attributes. Based on our analysis of swapping layers and channels, we have found that there are different combinations of layers or channels that can result in the best identity disentanglement score. 
This is perhaps not surprising since different people have different characterizing features and the identity disentanglement score combines the score over all attributes.} 

\revAA{}{In this appendix, we look closer at the correlation between the change in identity distance and the changes in individual attributes for a number of example transformations. Here, we chose to swap layers (5,6,7); i.e., the consecutive layers that were found to most effectively shift the identity to the target while maintaining a high disentanglement score.  Of particular interest is the correlation in the change in the identity distance (L2/Euclidean distance) and the changes in the attribute score (which is the confidence value of the attribute classifier $AttrNet$). Here, we measure the change of the individual attribute score using the L1 distance (as the single score of an attribute is a scalar value).} 

\revBB{}{Fig.~\ref{fig:id_corr} shows the Pearson correlation of the change in identity with all 40 attributes of CelebA dataset as seen for this particular example experiment. The six attributes with the highest correlation to the identity were: ``Wearing Lipstick", ``Male", ``Heavy Makeup", ``Attractive", ``Bushy Eyebrows", and ``High Cheekbones".  While some of these attributes (e.g., ``Wearing Lipstick" highest correlated to identity here) may appear non-intuitive at first, most of these top-ranked attributes help distinguish the gender of a client (e.g., ``Wearing Lipstick" and ``Heavy Makeup" may traditionally be more likely associated with females, whereas ``Male" and ``Bushy Eyebrows" may be better at indicating that a person is more likely to be male).
These clues are in line with our observation that gender is highly correlated with the identity and one of the attributes most \revCC{disentangle from}{entangled to} the identity.} 

\revAA{}{Some other attributes that also have some degree of correlation to the identity include ``Young", ``Wearing Necktie", ``Black Hair", ``Big Nose", etc. In contrast, attributes such as ``Mustache", ``Eyeglasses", ``Blond Hair" have a low or inverse correlation to the identity.}

\end{document}